\documentclass{article}

\usepackage{microtype}
\usepackage{graphicx}
\usepackage{subcaption}
\usepackage{booktabs} 

\usepackage{hyperref}
\usepackage{url}
\usepackage{xcolor} 
\usepackage[table]{xcolor}
\usepackage{colortbl} 
\usepackage{wrapfig}
\usepackage{multirow} 
\usepackage{arydshln} 
\usepackage{xspace}

\usepackage[accepted]{icml2026}

\usepackage{amsmath}
\usepackage{amssymb}
\usepackage{mathtools}
\usepackage{amsthm}

\usepackage[capitalize,noabbrev]{cleveref}

\theoremstyle{plain}

\theoremstyle{definition}

\theoremstyle{remark}

\definecolor{cred}{HTML}{FF6B6B}
\definecolor{cyellow}{HTML}{FEC260}
\definecolor{cgreen}{HTML}{70AD47}
\definecolor{cblue}{HTML}{4D96FF}
\definecolor{cpurple}{HTML}{2A0944}
\definecolor{ggray}{RGB}{127,127,127}
\definecolor{aliceblue}{rgb}{0.94, 0.97, 1.0}

\newcommand{\myparagraph}[1]{\textbf{#1}\hspace{1.8ex}}

\newcommand{\huggingface}{\raisebox{-1.5pt}{\includegraphics[height=1.05em]{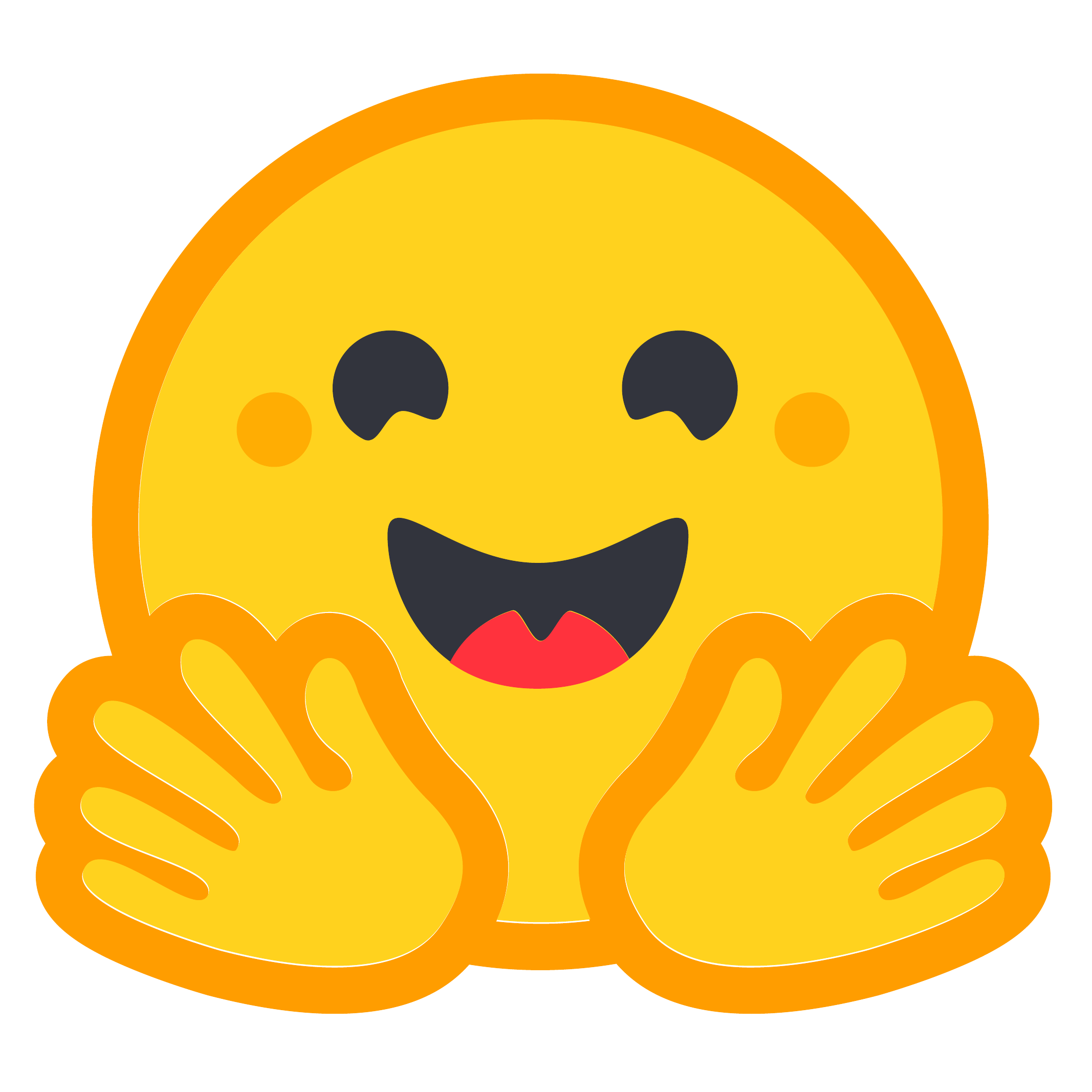}}\xspace}
\newcommand{\hfdataset}{\raisebox{-1.5pt}{\includegraphics[height=1.05em]{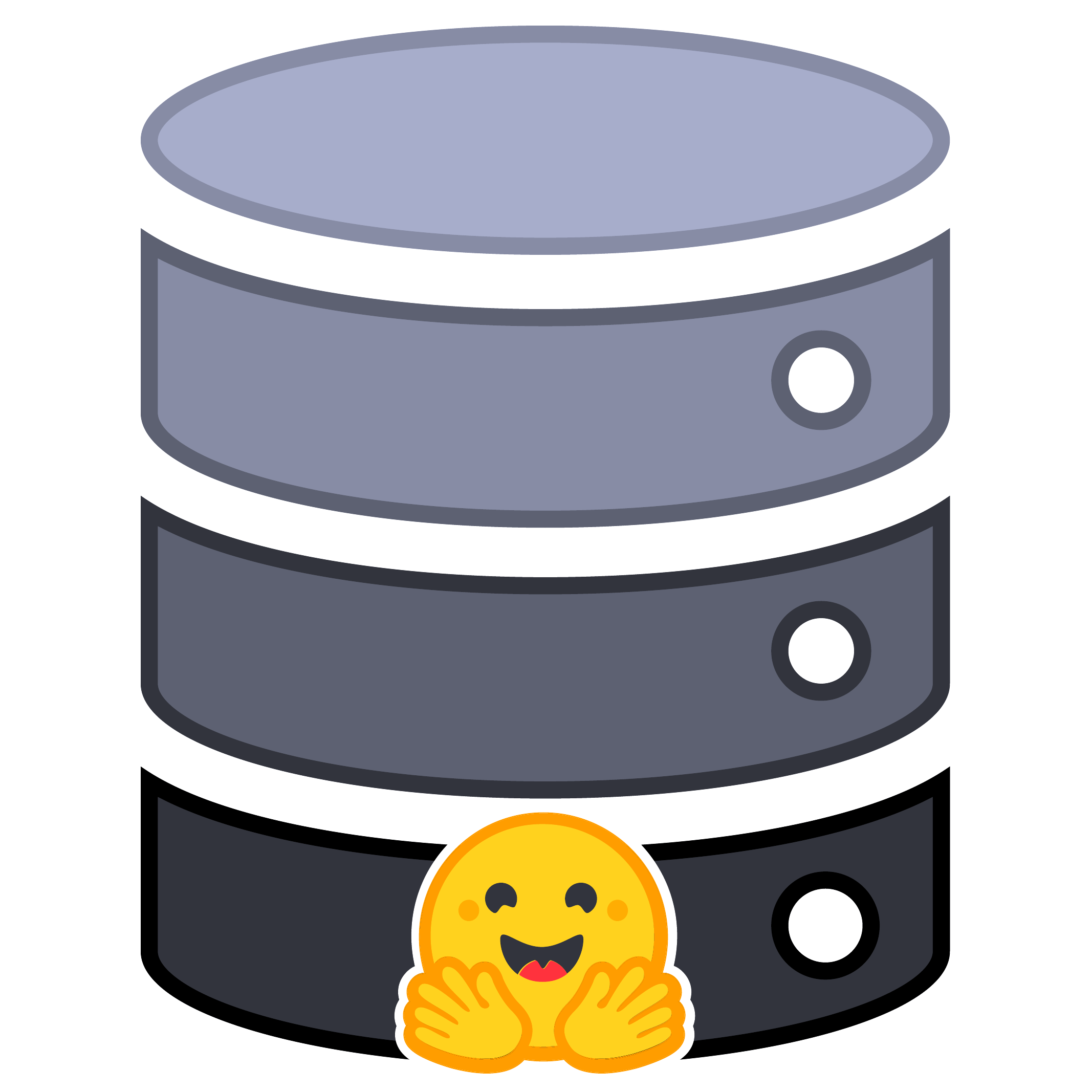}}\xspace}
\newcommand{\github}{\raisebox{-1.5pt}{\includegraphics[height=1.05em]{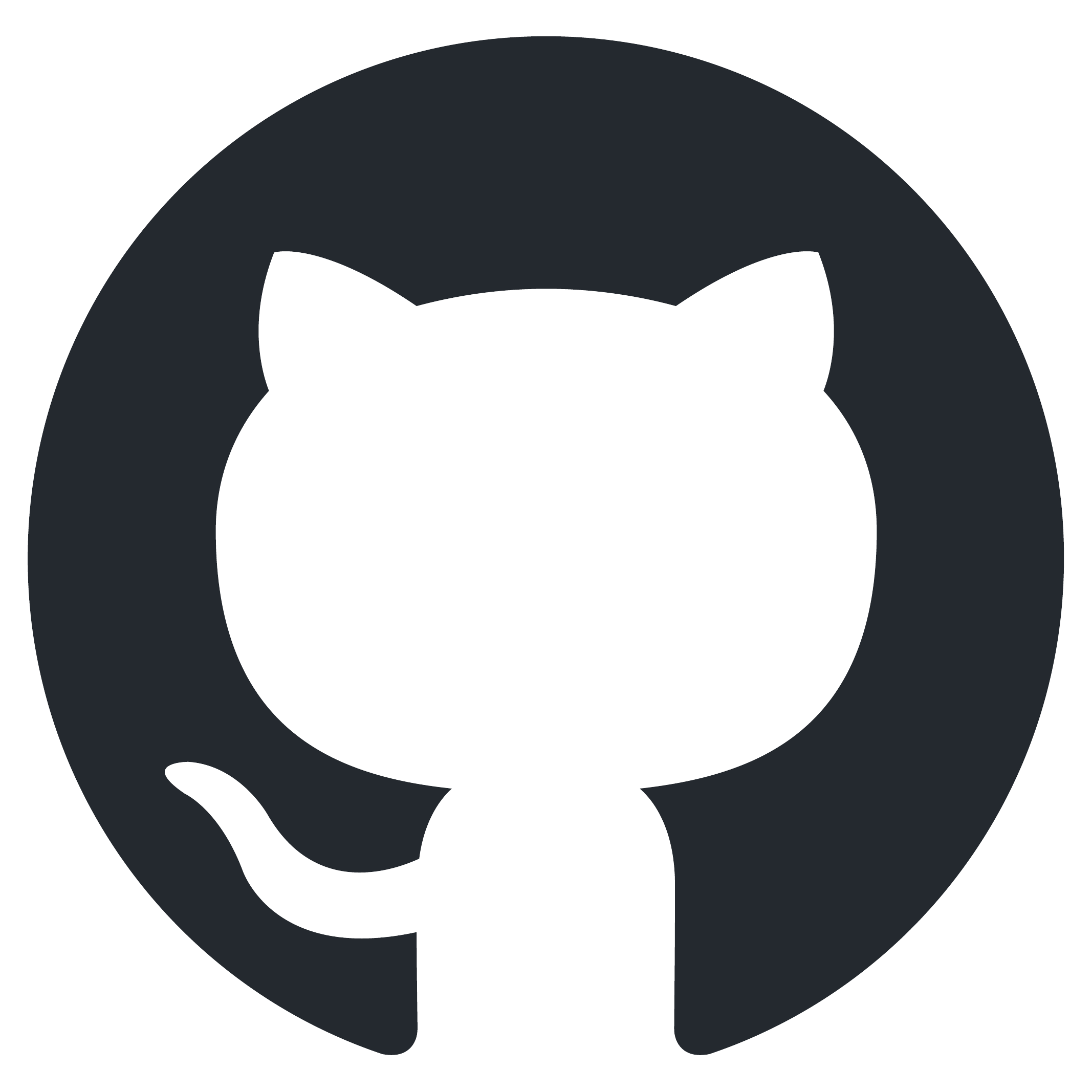}}\xspace}
\newcommand{\project}{\raisebox{0pt}{\includegraphics[height=1.0em]{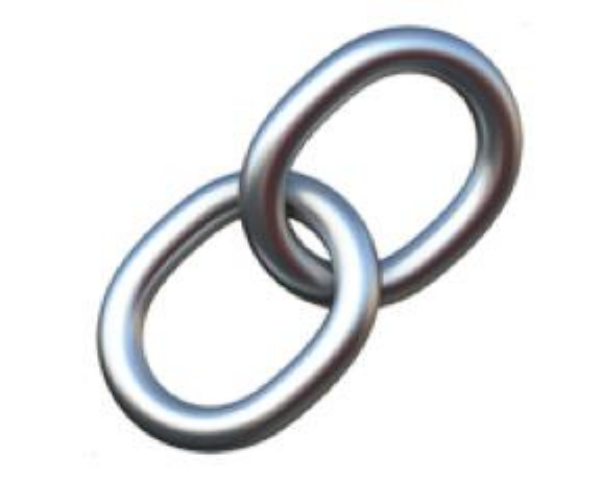}}\xspace}
\newcommand{\leaderboard}{\raisebox{0pt}{\includegraphics[height=1.0em]{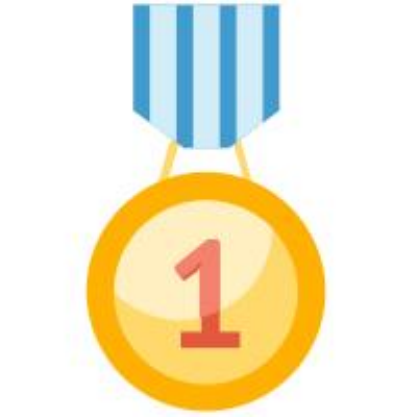}}\xspace}

\newcommand{\think}[1]{\textcolor{blue}{\texttt{<think>}} #1 \textcolor{blue}{\texttt{</think>}}}
\newcommand{\search}[1]{\textcolor{cyan}{\texttt{<search>}} #1 \textcolor{cyan}{\texttt{</search>}}}

\newcommand{\answer}[1]{\textcolor{purple}{\texttt{<answer>}} #1 \textcolor{purple}{\texttt{</answer>}}}
\newcommand{\judge}[1]{\textcolor{green}{\texttt{<judge>}} #1 \textcolor{green}{\texttt{</judge>}}}

\usepackage[textsize=tiny]{todonotes}

\icmltitlerunning{ReSeek: Self-Correcting Search Agents with Instructive Rewards}

\begin{document}

\twocolumn[
  \icmltitle{ReSeek: A Self-Correcting Framework for Search Agents with Instructive Rewards}



  \icmlsetsymbol{equal}{*}

  \begin{icmlauthorlist}
    \icmlauthor{Shiyu Li}{tencent}
    \icmlauthor{Yifan Wang}{tencent,tsinghua}
    \icmlauthor{Peiming Li}{tencent}
    \icmlauthor{Zheng Wei}{tencent}
    \icmlauthor{Yang Tang}{tencent}
  \end{icmlauthorlist}

  \icmlaffiliation{tencent}{Basic Algorithm Center, PCG, Tencent, Shenzhen, China}
  \icmlaffiliation{tsinghua}{Tsinghua Shenzhen International Graduate School, Tsinghua University, Shenzhen, China}

  \icmlcorrespondingauthor{Yang Tang}{ethanntang@tencent.com}
  \begin{center}
    \small
    \project\href{https://tencentbac.github.io/ReSeek/}{Homepage}\quad
    \github\href{https://github.com/TencentBAC/ReSeek}{Code}\quad
    \leaderboard\href{https://huggingface.co/spaces/TencentBAC/SearchAgent_Leaderboard}{Leaderboard}\quad
    \hfdataset\href{https://huggingface.co/datasets/TencentBAC/FictionalHot}{Dataset}\quad
    \huggingface\href{https://huggingface.co/TencentBAC/ReSeek-qwen2.5-3b-em-grpo}{Model}
  \end{center}
  \icmlkeywords{Machine Learning, ICML}

  \vskip 0.3in
]



\printAffiliationsAndNotice{}  

\begin{abstract}
Search agents powered by Large Language Models (LLMs) have demonstrated significant potential in tackling knowledge-intensive tasks. 
Reinforcement learning (RL) has emerged as a powerful paradigm for training these agents to perform complex, multi-step reasoning. 
However, prior RL-based methods often rely on sparse or rule-based rewards, which can lead agents to commit to suboptimal or erroneous reasoning paths without the ability to recover. 
To address these limitations, we propose \textbf{ReSeek}, a self-correcting framework enabling search agents to recover from erroneous search paths during an episode.
By invoking a special \textbf{JUDGE} action, the agent can judge the information and re-plan its search strategy. 
To guide this process, we design a dense, instructive process reward function, which combines an answer correctness reward for task completion with a self-correction reward that trains the agent to judge the utility of retrieved information.  
Additionally, to mitigate the risk of data contamination in existing datasets, we introduce \textbf{FictionalHot}, a contamination-resistant benchmark requiring complex reasoning. 
Experiments show ReSeek significantly outperforms SOTA baselines in task success and path faithfulness.
Our code and dataset are available at \url{https://github.com/TencentBAC/ReSeek}. 

\end{abstract}

\section{Introduction}

\begin{figure}[!t]
\begin{center}
\vspace{-.5em}
\centerline{\includegraphics[width=1.0\columnwidth]{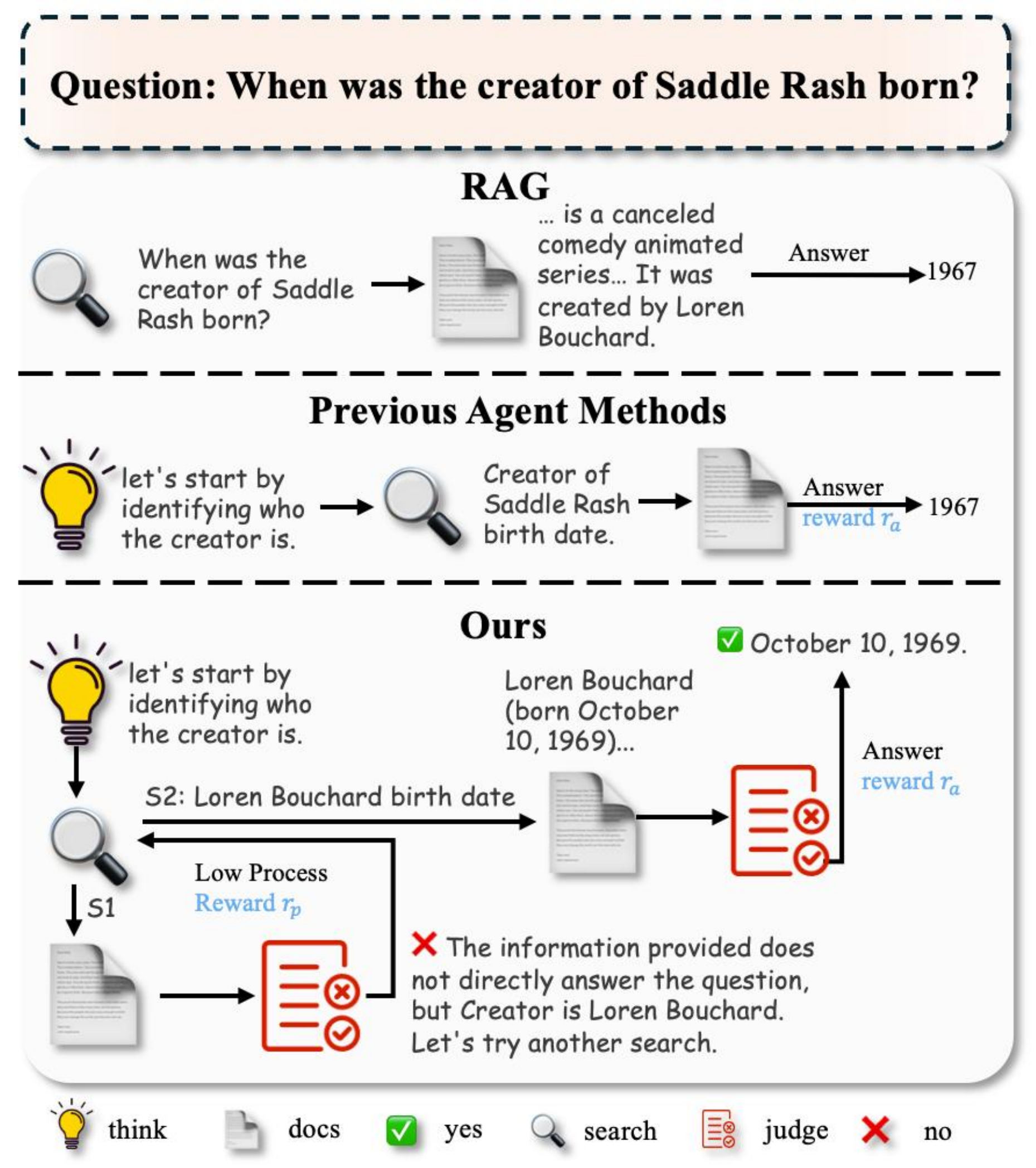}}
\caption{\textbf{A comparison of reasoning processes on a multi-hop question about an obscure entity.} Standard RAG fails as it cannot perform sequential reasoning. Vanilla agent reasons sequentially but gets stuck on its initial path. In contrast, our agent demonstrates robust self-correction: it uses a low process reward $(r_p)$ to identify the unproductive intermediate step, triggers a \textsc{judge} action to revise its strategy, and successfully navigates to the correct answer. The full trace for this example is provided in Appendix~\ref{appendix_case2}.}
\label{fig:intuition}
\end{center}
\vskip -0.4in
\end{figure}

Large Language Models (LLMs) ~\cite{brown2020language,chatgpt,ouyang2022training,guo2025deepseek,yang2025qwen3} have demonstrated unprecedented capabilities in natural language understanding and generation, yet they are inherently limited by their static, pre-trained knowledge, which can be outdated or lead to factual hallucinations~\cite{borgeaud2022improving,zhao2024recommender,maleki2024ai,zhang2025llm}. Search-augmented agents, which empower LLMs to interact with external tools like search engines, have emerged as a powerful paradigm to overcome these limitations~\cite{li2025webthinker,zheng2025deepresearcher,jin2025search,luo2025large}. By dynamically retrieving and reasoning over up-to-date information, these agents can tackle complex tasks that are beyond the reach of standalone LLMs.

Despite their promise, prevailing methods often falter in complex reasoning scenarios. 
Early approaches like Retrieval-Augmented Generation (RAG) are limited to a single retrieve-then-generate cycle, lacking the capability for error correction. 
More advanced RL-trained agents learn policies over action sequences, but their sparse or rule-based rewards (e.g., final answer correctness) provide insufficient guidance for intermediate steps. 
Consequently, an early mistake (such as a misleading search query) can cascade into an incorrect answer, as agents lack the ability to self-assess and recover. 
As shown in Figure~\ref{fig:intuition}, both RAG and agent methods~\cite{jin2025search} query for ``creator of Saddle Rash.'' 
The retrieved documents center almost on describing the show itself and contain no information about the director’s birth date, ultimately leading to an unsuccessful response.

To address these limitations, we propose ReSeek, a self-correcting framework for training search agents, centered on a special \textsc{judge} action. 
This action enables the agent to pause, evaluate the utility of retrieved information, and dynamically adapt its strategy when the current path proves unproductive.
As illustrated in Figure~\ref{fig:intuition}, the initial search revealed the creator's name (Loren Bouchard) but not the birth date. 
The \textsc{judge} action identified this gap and evaluated the retrieved information as useful yet insufficient. 
Consequently, the agent adapted its strategy by formulating a follow-up query ``Loren Bouchard birth date" and successfully retrieved the correct answer.
This entire process is guided by a \textbf{dense and instructive reward function} that combines an answer correctness reward for task completion with a self-correction reward for judging evidence utility, providing fine-grained, step-by-step feedback.

Another challenge lies in evaluation. Many existing datasets for knowledge-intensive tasks are at risk of data contamination. Their contents may overlap with LLM training corpora, causing high performance to reflect memorization rather than genuine reasoning.
To address this, we first introduce \textbf{FictionalHot}, a new benchmark composed of questions about fictional entities. 
Its design inherently mitigates contamination risk, forcing agents to rely on search results. 
Our contributions are summarized as follows:
\begin{itemize}
    
    
    
    \item We propose ReSeek, a novel framework enabling search agents to self-correct and dynamically adapt their search strategy.
    
    \item We design a dense, process-based reward function that combines an answer correctness reward with a self-correction reward for judging evidence utility.
    
    \item We introduce FictionalHot, a contamination-resistant benchmark for fair and challenging evaluation of search agents' reasoning abilities.
    
    \item Experiments show ReSeek outperforms SOTA baselines in performance and reasoning path faithfulness.
\end{itemize}

\section{Related Work}
\subsection{RAG and Search Agents}
To address knowledge limitations and hallucinations of LLMs~\cite{Zhang2023SirensSI, ji2023survey, bang2023multitask}, research has focused on integrating external knowledge. RAG frameworks~\cite{lewis2020retrieval, guu2020realm, karpukhin2020dense} enhance LLM responses by retrieving relevant documents, but their single-step cycle limits efficacy on multi-step reasoning tasks~\cite{jiang2023active, asai2023self}. To address this limitation, advanced Search Agents have been developed~\cite{shinn2023reflexion}, which decompose complex tasks by interleaving search and reasoning steps~\cite{yao2022react}. Works such as WebThinker~\cite{li2025webthinker}, DeepResearcher~\cite{zheng2025deepresearcher, jin2025search}, Search-o1~\cite{luo2024search}, and ZeroSearch~\cite{gao2024zerosearch} significantly improve performance on knowledge-intensive tasks by empowering LLMs to plan and execute multi-step strategies. For contamination-aware evaluation, MDBench~\cite{2025mdbench} constructs a synthetic multi-document reasoning benchmark from structured knowledge. Our FictionalHot reduces contamination at the source via fictional entity replacement.

\subsection{Search with Reinforcement Learning}
 By modeling the search process as a Markov Decision Process, Reinforcement Learning z(RL) can optimize an agent's policy to maximize cumulative rewards. Recent works such as Search-R1~\cite{jin2025search}, ToolRL~\cite{qin2024toolrl}, and others~\cite{deng2023agent, liu2023agent} have successfully applied RL to train LLMs to use search tools. A key challenge in this paradigm, however, lies in the design of the reward function. Many existing methods rely on sparse, outcome-based rewards (\emph{e.g.}, correctness of the answer)~\cite{jin2025search, deng2023agent}. While straightforward to implement, such rewards provide poor credit assignment for intermediate steps, offering little guidance for navigating complex reasoning paths~\cite{ouyang2022training}. Consequently, research has shifted towards denser guidance through process supervision and self-correction mechanisms~\cite{chen2023fireact}. For instance, Backtracking Correction~\cite{ICLR2025_80790082} refines the reasoning chain in a post-hoc manner, optimizing backwards from a completed trajectory. S2R~\cite{ma2025s2r} trains self-verification for mathematical reasoning. AgentPRM~\cite{xi2026agentprm} trains a separate process reward model for general interactive agents, while ReSeek uses a search-specific reranker reward and embeds \textsc{judge} directly as an internal action. In contrast to these prior directions, ReSeek enables dynamic self-correction within search-augmented QA trajectories.

\begin{figure*}[h]
\begin{center}
\centerline{\includegraphics[width=2.0\columnwidth]{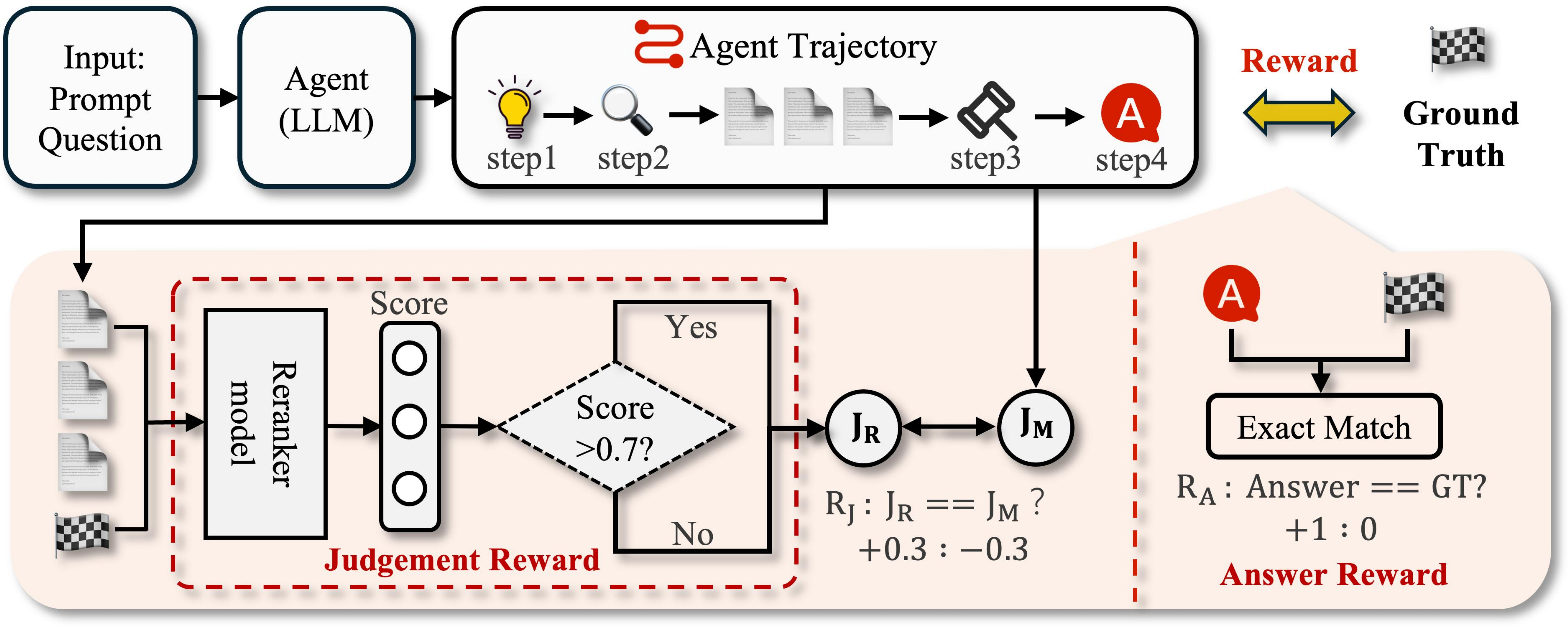}}
\caption{\textbf{Training the agent's self-evaluation capability.} We train the agent via policy optimization to master the \textsc{judge} action. A reward signal is generated by comparing the agent's judgment against an ``ideal" one, which is determined by the rerank score between the current search observation and the GT answer. This reward guides the policy to learn effective self-correction.}
\label{fig:framework}
\end{center}
\end{figure*}

\section{Methodology}
In this section, we present a framework for training LLM agents on complex tasks. Our approach is centered on enhancing the agent's decision-making capabilities through a novel reinforcement learning paradigm. We begin by formally defining the problem.

\subsection{Problem Formulation}
We formulate the agent's interaction as a Markov Decision Process (MDP). The agent is represented by a policy $\pi_\theta$ parameterized by an LLM. At each step $t$, given the state $s_t$, the policy generates an action $a_t \in \mathcal{A}$. 

To enable self-correction, we define the action space $\mathcal{A}$ to encompass not only standard tool invocations but also evaluation steps (detailed as the \textsc{judge} mechanism in Sec.~\ref{sec:judge}). We optimize $\pi_\theta$ against a reference policy $\pi_{\text{ref}}$ using a Group Relative Policy Optimization (GRPO) based objective:
\begin{equation}
\small
\max_{\pi_\theta} \mathbb{E}_{x \sim \mathcal{D}, y \sim \pi_\theta(\cdot|x) }[R(x, y)] - \beta D_{KL} [\pi_\theta(y | x) || \pi_{\text{ref}}(y | x)]
\end{equation}
Here, $y$ represents the generated trajectory for a problem instance $x$ from dataset $\mathcal{D}$. $R(x, y)$ aggregates rewards from task completion and intermediate evaluations, while the $\beta$-weighted KL term regularizes $\pi_\theta$ towards $\pi_{\text{ref}}$ for stability.

\subsection{JUDGE Action}\label{sec:judge}
A core challenge for search agents is reliability; a single error can derail an entire task. To operate robustly, agents require meta-cognitive awareness to assess the utility of intermediate steps. To achieve this, we introduce the \textsc{judge} action, a mechanism designed to serve as the agent's instrument for introspection. The \textsc{judge} action converts the agent's reasoning from a static, linear chain into a \emph{dynamic self-correction loop}.

Instead of physically removing uninformative steps (which might lead to repeating the same errors), we employ a \textbf{soft removal} strategy. The agent executes a judgment action to generate a label $j_t$ for the current observation $o_t$. This label is appended to the interaction history, serving as a signal that guides the policy to semantically attend to or disregard the preceding information.

We formalize this process as follows. Let $\tau_{t-1}$ denote the interaction trajectory up to step $t-1$. The policy conditions its next action $a_{t+1}$ on the updated context $\mathcal{C}_t$:
\begin{equation}
a_{t+1} \sim \pi(\cdot | \mathcal{C}_t) \quad \text{where} \quad \mathcal{C}_t = \tau_{t-1} \oplus a_t \oplus o_t \oplus j_t
\end{equation}
Here, $\oplus$ denotes concatenation. In this formulation, the judgment $j_t \in \{\text{'good'}, \text{'bad'}\}$ acts as a \textbf{cognitive filter}. A positive judgment reinforces the evidence $o_t$, while a negative judgment marks that $o_t$ is unhelpful, effectively ``blocking" the error path to prevent repeated mistakes.

\begin{table*}[h]
    \caption{\textbf{Self-corrective agent prompt.} Following each search, the model executes \texttt{<judge>} to determine if it should search again or answer. The Question placeholder is replaced at runtime with the current query.}\label{tab:instruction}
    \centering
    \resizebox{\textwidth}{!}{
    \begin{tabular} {p{17cm}}
        \toprule[1.25pt]
        Answer the given question step by step.
     Instructions: \\
     1. First, conduct reasoning inside \think{and} tags whenever you receive new information. \\
     2. If you need external knowledge, you can search using \search{query}. \\
     3. When you receive search results, evaluate their usefulness and put your judgment inside \judge{Yes} or \judge{No} tags. \\
     4. Based on your judgment, follow these strict rules: \\
     \quad a. If the information is useful AND you now have sufficient information to provide a complete final answer, proceed directly to step 5.\\
     \quad b. If the information is useful BUT you still need more details, you MUST search again with \search{...}\\
     \quad c. If the information is not useful, you MUST search again with \search{...}. You MUST NOT provide an answer in this case.\\
     5. Provide your final answer in \answer{...} tags. The \texttt{<answer>} tag marks the end of the task. After providing the \texttt{<answer>}, you MUST stop and generate no further text.\\
     Question: [question] \\
     \bottomrule[1.25pt]
    \end{tabular}}
\end{table*}

\subsection{Reward Function}
Our reward function is composed of two distinct signals: a rule-based reward for task completion and a dense intermediate reward for self-correction.

First, we define the standard objective for the task. The Correctness Reward is a rule-based signal assigned only after the final \textsc{answer} action. It is based on an Exact Match (EM) with the ground truth $A_g$ and prediction $A_p$:
\begin{equation}
R_{\text{answer}} = EM(A_p,A_g)
\end{equation}

While this provides the ultimate objective, relying solely on this sparse and delayed signal makes it difficult for the agent to learn the nuanced skill of step-by-step self-assessment.

To address the sparsity of the terminal reward, we introduce a dense, intermediate reward signal specifically designed to cultivate the agent's self-correction faculty. The core idea is to reward the agent for making judgments that align with an objective evaluation of the retrieved information.

We first establish an objective ``Ideal Judgment," $j^*_t$, for each observation. This ideal is based on the principle that retrieved information (observation) is useful if it exhibits high semantic similarity to the ground truth. We formalize this as:
\begin{equation}
j^*_t =
\begin{cases}
\text{`Yes'} & \text{if }\mathrm{score}(o_t,\mathrm{gt})>0.7 \\
\text{`No'} & \text{otherwise}
\end{cases}
\end{equation}
Here, $o_t$ represents the retrieved information (observation) at step $t$. The score function is implemented using bge-reranker-large~\cite{bge_embedding}, which computes the semantic relevance between the observation $o_t$ and the ground-truth answer GT. The threshold of 0.7 was determined empirically (Sec~\ref{sec:Threshold}), as it provided a separation between useful and non-useful information. 

With the ideal judgment established, we define the self-correction reward $R_{\text{judge}}$ which is given whenever the agent performs a \textsc{judge} action:
\begin{equation}
R_{\text{judge}}(j_t, j^*_t) =
\begin{cases}
R_{match} & \text{if } j_t = j^*_t \\
R_{mismatch} & \text{if } j_t \neq j^*_t
\end{cases}
\end{equation}
Here, $j_t$ is the agent's judgment. $R_{\text{match}}$ and $R_{\text{mismatch}}$ is the reward value.  
The $R_{\text{match}}$ provides a reward of 0.3 for both types of correct judgments: correctly identifying useful information ($j_t=j^*_t=\text{Yes}$) and discarding useless information ($j_t=j^*_t=\text{No}$).   
The $R_{\text{mismatch}}$ implements asymmetric penalties: correctly accepting useless information incurs a large penalty of -0.6, while discarding useful information incurs a smaller penalty of -0.3.

Finally, the complete training objective is to maximize the total reward over the trajectory $\tau$:
\begin{equation}
R(\tau) = \sum_{t=1}^{T} \gamma^{t-1} r_t
\end{equation}
where $r_t$ is defined conditionally based on the temporal stage of the process, specifically corresponding to the judgment reward $R_{\text{judge}}$ during the intermediate retrieval steps (where $t < T$) and transitioning to the answer reward $R_{\text{answer}}$ at the final terminal step (where $t = T$). This combination ensures the agent is guided by both immediate feedback and the ultimate answer.





\subsection{Structured Prompting}

\begin{table*}[h]
\caption{\textbf{Variety in Experimental Setups of Prior Works.} LJ means the metric of LLM-as-a-judge. This Table highlights the diversity in test sets, training sets, corpus, and evaluation metrics.}\label{tab:setup_comparison}
\scriptsize
\centering
\resizebox{\textwidth}{!}{
\begin{tabular}{lcccc}
\toprule[1.25pt]
\textbf{Methods} & \textbf{Test Sets} & \textbf{Training Sets} & \textbf{Corpus} & \textbf{Metrics} \\
\midrule
Search-o1~\cite{luo2024search}     & Set A & None & Internet & Exact Match \\
R1-Searcher~\cite{songR1SearcherIncentivizingSearch2025}   & Set B & HotpotQA, 2wiki & 2019-wiki & Cover Exact Match, LJ \\
Search-R1~\cite{jin2025search}     & Set A & NQ, TriviaQA & 2018-wiki & Exact Match \\
ReSearch~\cite{chen2025learning}     & Set B & Musique & 2018-wiki & Exact Match, LJ \\
R1-Searcher++~\cite{songR1SearcherIncentivizingDynamic2025} & Set B & HotpotQA, 2wiki & 2019-wiki & F1, LJ \\
Deepresearcher~\cite{zheng2025deepresearcher}  & Set A & NQ, TriviaQA, HotpotQA, 2wiki & Internet & F1, MBE \\
ZeroSearch~\cite{sunZeroSearchIncentivizeSearch2025}    & Set A & NQ, TriviaQA & Internet & Substring Exact Match \\
\bottomrule[1.25pt]
\end{tabular}}
\end{table*}

The self-correction mechanism fundamentally relies on the agent's ability to generate structured trajectories of actions and observations. 
To this end, we designed the structured prompt in Table \ref{tab:instruction} to enforce a strict, self-corrective reasoning prompt. 
It achieves this by incorporating two key architectural constraints: first, a mandatory \texttt{<judge>} action creates an explicit self-assessment checkpoint after every information retrieval step. 
Second, strict conditional rules (Rule 4) make the agent's subsequent actions contingent on the judgment's outcome. 
The explicitness of these rules ensures immediate and reliable trajectory generation, even from an untrained LLM.

\subsection{FictionalHot Benchmark}\label{sec:fictionalHot}
The lack of standardized experimental settings hinders the robust evaluation of search agents. Table~\ref{tab:setup_comparison} collates representative setups, highlighting variation in (i) corpora—from static Wikipedia snapshots (e.g., 2018, 2019) to the non-reproducible, live Internet; (ii) test sets—either a broad Set A (NQ\citep{Kwiatkowski_Palomaki_Redfield_Collins_Parikh_Alberti_Epstein_Polosukhin_Devlin_Lee_e2019}, TriviaQA\citep{Joshi_Choi_Weld_Zettlemoyer_2017}, PopQA\citep{mallenWhenNotTrust2022}, HotpotQA\citep{yangHotpotQADatasetDiverse2018}, 2Wiki\citep{hoConstructingMultihopQA2020}, Musique\citep{trivediMuSiQueMultihopQuestions2022}, Bamboogle\citep{pressMeasuringNarrowingCompositionality2022}) or a focused multi-hop Set B (HotpotQA, 2Wiki, Musique, Bamboogle); (iii) training regimes—ranging from no training to single or multi-dataset setups (e.g., HotpotQA, 2Wiki, NQ, TriviaQA); and (iv) metrics—spanning Exact Match and F1 to model-based judgments such as LLM-as-a-judge (LJ). This diversity reflects rapid progress but prevents direct comparison.


Beyond standardization, a deeper challenge is data contamination, where high scores on existing benchmarks can reflect memorized pretraining knowledge rather than genuine procedural reasoning. To address both issues, we introduce FictionalHot, a closed-world benchmark built on a synthetic corpus of fictitious entities. This design substantially reduces pretraining bias, forcing agents to rely primarily on procedural reasoning for a rigorous evaluation.

\begin{figure}[h]
\begin{center}
\centerline{\includegraphics[width=1.0\columnwidth]{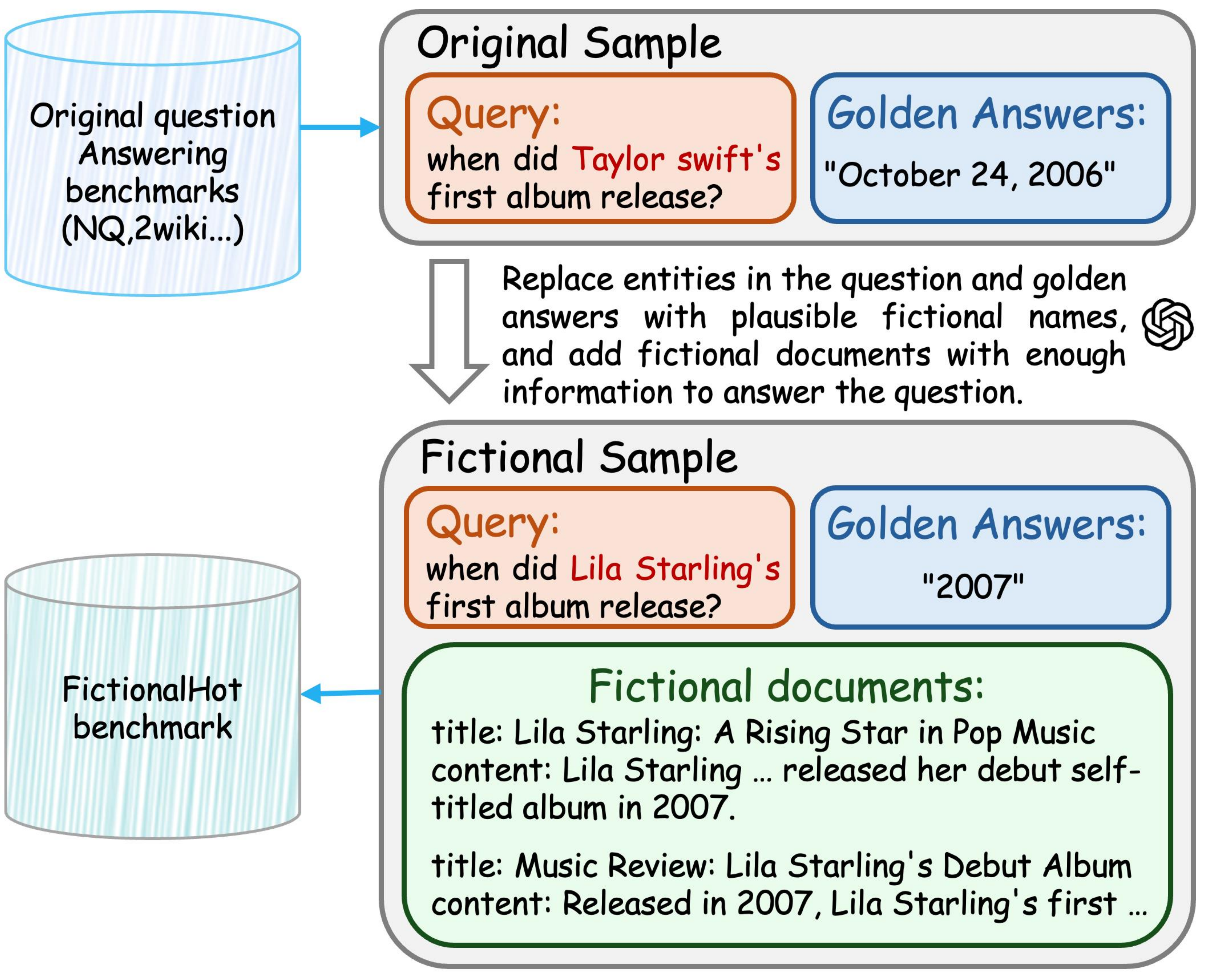}}
\caption{\textbf{FictionalHot construction process:} transforming a real-world question answer sample into a fictional sample with fictional question and documents.}
\label{fig:datapipe}
\vspace{-20pt}
\end{center}
\end{figure}
The construction of FictionalHot follows a three-step pipeline, as illustrated in the Figure~\ref{fig:datapipe}. The FictionalHot dataset contains a total of 5,116 questions. First, these were generated by drawing a 10\% random sample of seed questions from a pool of 51,588 samples across six major benchmarks (Bamboogle was excluded due to its limited size of 125 samples).
Next, these questions are paraphrased by GPT-5. 
This core step replaces real-world entities with plausible, fictional ones (e.g., `Taylor Swift' becomes `Lila Starling') while preserving the original question's reasoning structure. 
Crucially, GPT-5 also generates new Wikipedia-style documents for these fictional entities, creating a new, self-contained fact (e.g., setting the album release to `2007') that serves as the basis for the new golden answer. 
Finally, to create the closed-world corpus, these synthetic samples are inserted into the 2018 Wikipedia corpus. 

\section{Experiments}

\begin{table*}[t]
    \caption{\textbf{Overall performance comparison of ReSeek against baselines across a comprehensive suite of QA benchmarks.} Results are reported in Exact Match (EM) for both Qwen2.5-7B and 3B backbones. The best results are in \textbf{bold}.}\label{tab:main}
    \centering
    \resizebox{\textwidth}{!}{
    \begin{tabular}{lccccccccc}
        \toprule[1.25pt]
        \textbf{Methods} & \multicolumn{3}{c}{\textbf{General QA}} & \multicolumn{4}{c}{\textbf{Multi-Hop QA}} \\
        \cmidrule(r){2-4} \cmidrule(r){5-10}
         & \textbf{NQ} & \textbf{TriviaQA} & \textbf{PopQA} & \textbf{HotpotQA} & \textbf{2wiki} & \textbf{Musique} & \textbf{Bamboogle} & \textbf{FictionalHot} & \textbf{Avg.} \\
        \midrule
        \multicolumn{8}{l}{\textbf{Qwen2.5-7b-Instruct}} \\
        Direct Inference & 0.134 & 0.408 & 0.140 & 0.183 & 0.250 & 0.031 & 0.120 & 0.001 & 0.158  \\
        CoT & 0.048 & 0.185 & 0.054 & 0.092 & 0.111 & 0.022 & 0.232 & 0.001 & 0.093\\
        RAG & 0.349 & 0.585 & 0.392 & 0.299 & 0.235 & 0.058 & 0.208 & 0.012 & 0.267\\
        R1~\cite{guo2025deepseek} & 0.270 & 0.537 & 0.199 & 0.237 & 0.292 & 0.072 & 0.293 & 0.003 & 0.238 \\
        Search-o1~\cite{luo2024search} & 0.151 & 0.443 & 0.131 & 0.187 & 0.176 & 0.058 & 0.296 & 0.020 & 0.183\\
        Search-R1~\cite{jin2025search} & 0.393 & 0.610 & 0.397 & 0.370 & \textbf{0.414} & 0.146 & 0.368 & 0.034 & 0.342 \\
        ZeroSearch~\cite{sunZeroSearchIncentivizeSearch2025} & 0.436 & \textbf{0.652} & 0.488 & 0.346 & 0.352 & 0.184 & 0.278 & 0.031 & 0.346 \\
        \rowcolor{aliceblue!60} ReSeek & \textbf{0.469} & 0.640 & \textbf{0.501} & \textbf{0.389} & 0.382 & \textbf{0.185} & \textbf{0.392} & \textbf{0.061} & \textbf{0.377} \\
        \midrule
        \multicolumn{8}{l}{\textbf{Qwen2.5-3b-Instruct}} \\
        Direct Inference & 0.106 & 0.288 & 0.108 & 0.149 & 0.244 & 0.020 & 0.024 & 0.001 & 0.118 \\
        CoT & 0.023 & 0.032 & 0.005 & 0.021 & 0.021 & 0.002 & 0.000 & 0.001 & 0.013 \\
        RAG & 0.348 & 0.544 & 0.387 & 0.255 & 0.226 & 0.047 & 0.080 & 0.008 & 0.237  \\
        R1~\cite{guo2025deepseek}  & 0.210 & 0.449 & 0.171 & 0.208 & 0.275 & 0.060 & 0.192 & 0.003 & 0.196  \\
        Search-o1~\cite{luo2024search} & 0.238 & 0.472 & 0.262 & 0.221 & 0.218 & 0.054 & 0.019 & 0.010 & 0.187 \\
        Search-R1~\cite{jin2025search} & 0.341 & 0.545 & 0.378 & 0.324 & 0.319 & 0.103  & 0.264  & 0.037 & 0.288 \\
        ZeroSearch~\cite{sunZeroSearchIncentivizeSearch2025} & 0.414 & \textbf{0.574} & \textbf{0.448} & 0.274 & 0.300 & 0.098 & 0.111 & 0.030 & 0.281 \\
        \rowcolor{aliceblue!60} ReSeek & \textbf{0.415} & 0.553 & 0.434 & \textbf{0.328} & 0.298 & \textbf{0.103} & \textbf{0.304} & \textbf{0.059} & \textbf{0.312} \\
        \bottomrule[1.25pt]
    \end{tabular}}
\end{table*}

\subsection{Experimental Setup}
\myparagraph{Evaluation Benchmarks.} 
We evaluate on 51,713 samples across two categories: 
(i) \textbf{Classic Open-domain QA:} Single-hop (PopQA, TriviaQA, NQ) and multi-hop (2Wiki, HotpotQA, Musique, Bamboogle) datasets.
(ii) \textbf{FictionalHot:} mitigates contamination via a closed-world, synthetic corpus.
To ensure fair comparison, we follow~\cite{jin2025search} and adopt Exact Match (EM) as the primary metric. A prediction is considered correct if its normalized string exactly 
matches any normalized reference answer. Normalization applies lowercasing, removes punctuation and articles, and collapses whitespace.

\noindent\myparagraph{Training and Evaluation Setup.}
Following Search-R1~\cite{jin2025search}, we fine-tune exclusively on a unified set of NQ and HotpotQA (169,615 pairs) from FlashRAG~\cite{FlashRAG}.
We use Qwen-2.5-3B-Instruct and 7B-Instruct~\cite{qwen2025qwen25technicalreport}.
At test time, retrieval top-k is set to $k=3$ with a maximum of $T=4$ tool-use turns per question. Experiments are run on 16 H20 GPUs; the search backend uses E5 embeddings\citep{wang2022text} on Wiki-18 corpus (wiki-18)~\cite{karpukhin2020dense}. GRPO is used as the default RL algorithm, and a detailed comparison with PPO is provided in Appendix~\ref{appendix_ppo}. Additional details are provided in Appendix~\ref{appendix_details}.

\noindent\myparagraph{Baselines.} 
We compare with four baseline families: Vanilla prompting (zero-shot direct answering and Chain-of-Thought with no external search)~\cite{karpukhin2020dense}; Single-pass RAG (retrieve once, then generate conditioned on the top‑k passages)~\cite{lewis2020retrieval}; Agentic search (multi-step search–reason loops such as ReAct-style planners, without RL tuning)~\cite{chung2024scaling, luo2024search}; and RL‑tuned policies~\cite{jin2025search,chen2025learning,sunZeroSearchIncentivizeSearch2025}.

\subsection{Main results}

We evaluate ReSeek across eight open-domain QA benchmarks spanning single- and multi-hop settings, using Qwen2.5-7B-instruct and Qwen2.5-3B-instruct backbones. 

\noindent\myparagraph{ReSeek achieves SOTA performance.} We evaluate ReSeek across eight open-domain QA benchmarks spanning single- and multi-hop settings. ReSeek attains the highest average accuracy across both backbones: 0.377 for 7B compared to 0.346 for ZeroSearch, and 0.312 for 3B compared to 0.281. 
It consistently excels on multi-hop benchmarks, notably HotpotQA and Bamboogle across both model scales, which highlights the benefits of our repeated-search with self-correction paradigm. On single-hop datasets, ZeroSearch performs competitively, which aligns with its design focus on direct retrieval.

\noindent\myparagraph{FictionalHot isolates reasoning ability from model scale and data leakage.} 
On FictionalHot, ReSeek scores 0.061 (7B) and 0.059 (3B). This consistent performance across scales indicates that FictionalHot effectively isolates reasoning ability from scale-correlated memorization.
In contrast, TriviaQA shows a large gap (0.408 for 7B vs. 0.288 for 3B), while Direct Inference on FictionalHot is near-zero (0.001). This pattern suggests training data overlap in TriviaQA, whereas FictionalHot's reduced-contamination design provides a cleaner measure of genuine reasoning.

\subsection{Ablative analysis}
\myparagraph{Ablation Study on Each Component.} 
We conduct an ablation study to evaluate the impact of the \textsc{judge} action and the dense intermediate utility reward ($R_{\text{judge}}$). As shown in Table \ref{tab:ablation}, enabling the \textsc{judge} action improves the average score from 0.288 to 0.297, demonstrating the benefit of adaptive retrieval termination. The dense utility reward ($R_{\text{judge}}$) further boosts the score to 0.312 (+7.7\%), confirming that dense signals are vital for reliable information assessment.
\begin{table*}[h]
\caption{\textbf{Ablation study of model components.} We report the Average Score to quantify the contribution of the \textsc{judge} action and the intermediate utility reward.}
\label{tab:ablation}
\centering
\resizebox{\textwidth}{!}{
\begin{tabular}{lccccccccc}
\toprule
\textbf{Component} & \multicolumn{3}{c}{\textbf{General QA}} & \multicolumn{4}{c}{\textbf{Multi-Hop QA}} \\
\cmidrule(r){2-4} \cmidrule(r){5-10} &
\textbf{NQ} & \textbf{TriviaQA} & \textbf{PopQA} & \textbf{HotpotQA} & \textbf{2wiki} & \textbf{Musique} & \textbf{Bamboogle} & \textbf{FictionalHot} & \textbf{Avg.} \\
\midrule
 $R_{\text{answer}}$ & 0.341 & 0.545 & 0.378 & 0.324 & 0.319 & 0.103  & 0.264  & 0.037 & 0.288 \\
+ \textsc{judge} Action & 0.370 & 0.550 & 0.405 & 0.324 & 0.293 & 0.103 & 0.275 & 0.052 & 0.297 \\
+ $R_{\text{judge}}$ (ReSeek) & \textbf{0.415} & 0.553 & 0.434 & \textbf{0.328} & 0.298 & \textbf{0.103} & \textbf{0.304} & \textbf{0.059} & \textbf{0.312} \\
\bottomrule
\end{tabular}}
\end{table*}
\begin{table*}[h]
\caption{\textbf{Ablation study on the reranker component of our reward function.} Our method, ReSeek, uses the BGE-Reranker. We compare it against variants with a different neural reranker (Qwen), a heuristic reranker (Regex-based), or no reranker at all (None).}
\label{tab:ablation_reranker}
\centering
\resizebox{\textwidth}{!}{
\begin{tabular}{lccccccccc}
\toprule[1.25pt]
\textbf{Methods} & \multicolumn{3}{c}{\textbf{General QA}} & \multicolumn{4}{c}{\textbf{Multi-Hop QA}} \\
\cmidrule(r){2-4} \cmidrule(r){5-10}
& \textbf{NQ} & \textbf{TriviaQA} & \textbf{PopQA} & \textbf{HotpotQA} & \textbf{2Wiki} & \textbf{Musique} & \textbf{Bamboogle} & \textbf{FictionalHot} & \textbf{Avg.} \\
\midrule
None (w/o Reranker) & 0.391 & 0.495 & 0.362 & 0.255 & 0.218 & 0.081 & 0.243 & 0.025 & 0.259 \\
Regex-based & 0.410 & 0.541 & 0.422 & 0.320 & 0.291 & 0.093 & 0.288 & 0.042 & 0.301 \\
Qwen-Reranker & 0.413 & \textbf{0.557} & 0.432 & 0.326 & \textbf{0.301} & 0.101 & 0.302 & 0.057 & 0.311 \\
\rowcolor{aliceblue!60}
ReSeek (Ours, w/ BGE) & \textbf{0.415} & 0.553 & \textbf{0.434} & \textbf{0.328} & 0.298 & \textbf{0.103} & \textbf{0.304} & \textbf{0.059} & \textbf{0.312} \\
\bottomrule[1.25pt]
\end{tabular}}
\end{table*}

\noindent\myparagraph{Ablation Study on the Reranker Component.}
To assess the overall effectiveness of our reward function and the specific choice of its reranker component, we compare ReSeek (BGE-Reranker) against three baselines: (i) None (no reranker), (ii) Qwen-Reranker~\cite{zhang2025qwen3}, and (iii) Regex-based. The Regex-based method extracts `Yes/No' judgments from the reasoning trace. It validates these by checking if the final answer appears in the retrieved text, applying rewards or penalties.
Table \ref{tab:ablation_reranker} shows a clear hierarchy: while Regex-based improves upon None, neural rerankers (BGE and Qwen) yield the largest gains, highlighting the superiority of semantic understanding over lexical matching for complex queries.

\noindent\myparagraph{Reranker-Only vs.\ RL-Trained Judgment.}
To separate the value of the reranker signal from RL training, we compare three variants on the 7B model (Table~\ref{tab:ablation_prompt}): (1)~\emph{Reranker-only intervention} applies bge-reranker-large scores directly at inference to determine good/bad labels and soft removal, without RL training; (2)~\emph{Prompt-only} adds ReSeek's \textsc{judge} prompt to Search-R1 without RL training; and (3)~full ReSeek with GRPO. Reranker-only improves Search-R1 by +1.2 avg EM, while full ReSeek adds another +2.3 avg EM, confirming that RL training teaches the agent to judge in context beyond raw reranker signals.

\begin{figure}[h]
    \centering
    \includegraphics[width=\columnwidth]{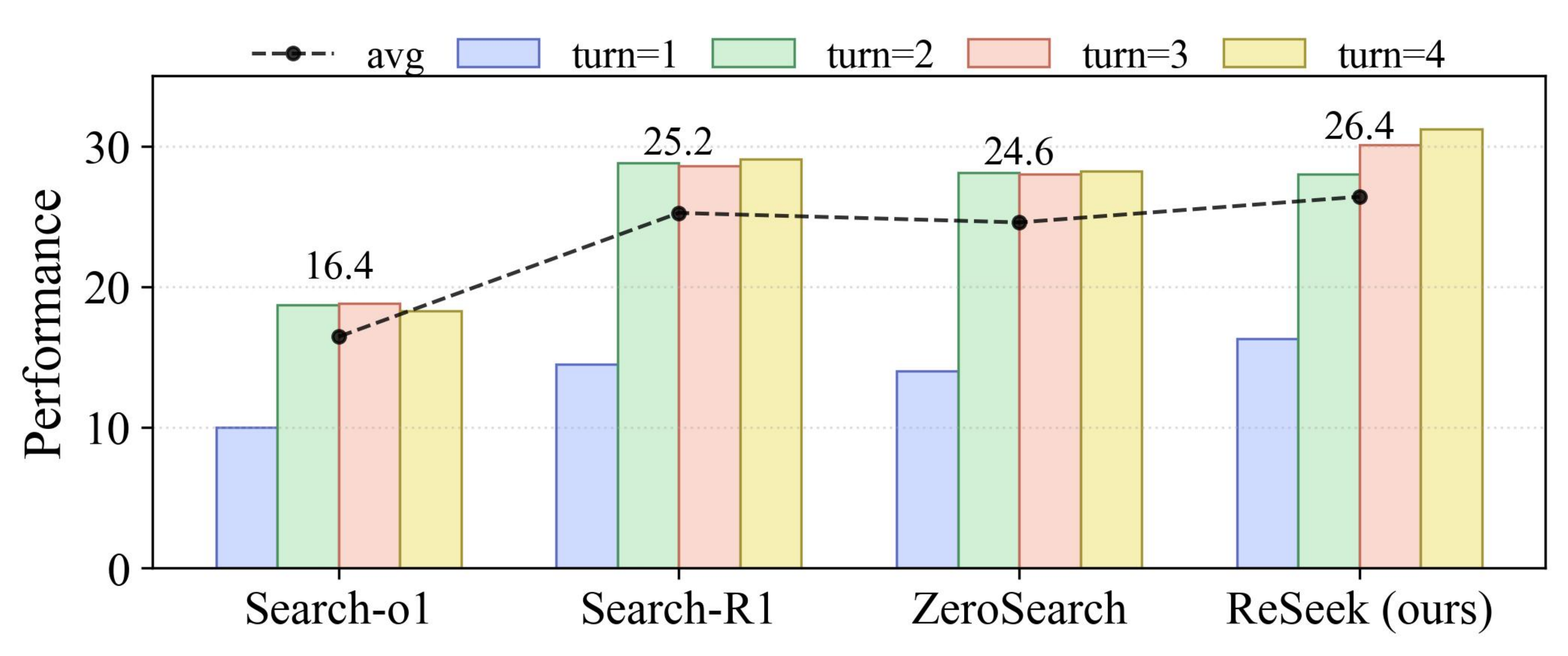}
    \caption{\textbf{Ablation study on the effect of the number of turns.} We evaluate methods with turn budgets from 1 to 4 using qwen2.5-3b-instruct, reporting the average performance across all datasets.}
    \label{fig:abl_turns}
\end{figure}

\noindent\myparagraph{Interaction Turns Study.} 
We perform an ablation over the number of turns to isolate the effect of the action budget and to test whether models can leverage iterative self-correction. 
Here, turns denotes the maximum number of actions the model may execute for a query. 
As shown in Figure~\ref{fig:abl_turns}, the baselines improve substantially from 1 to 2 turns but saturate thereafter, consistent with a simple search-then-answer workflow (typically one turn to search and one turn to answer).
In contrast, ReSeek improves monotonically up to four turns, confirming its ability to leverage extra steps for re-querying and plan refinement, converting larger budgets into genuine gains.

\begin{figure}[h]
    \centering
    \includegraphics[width=\columnwidth]{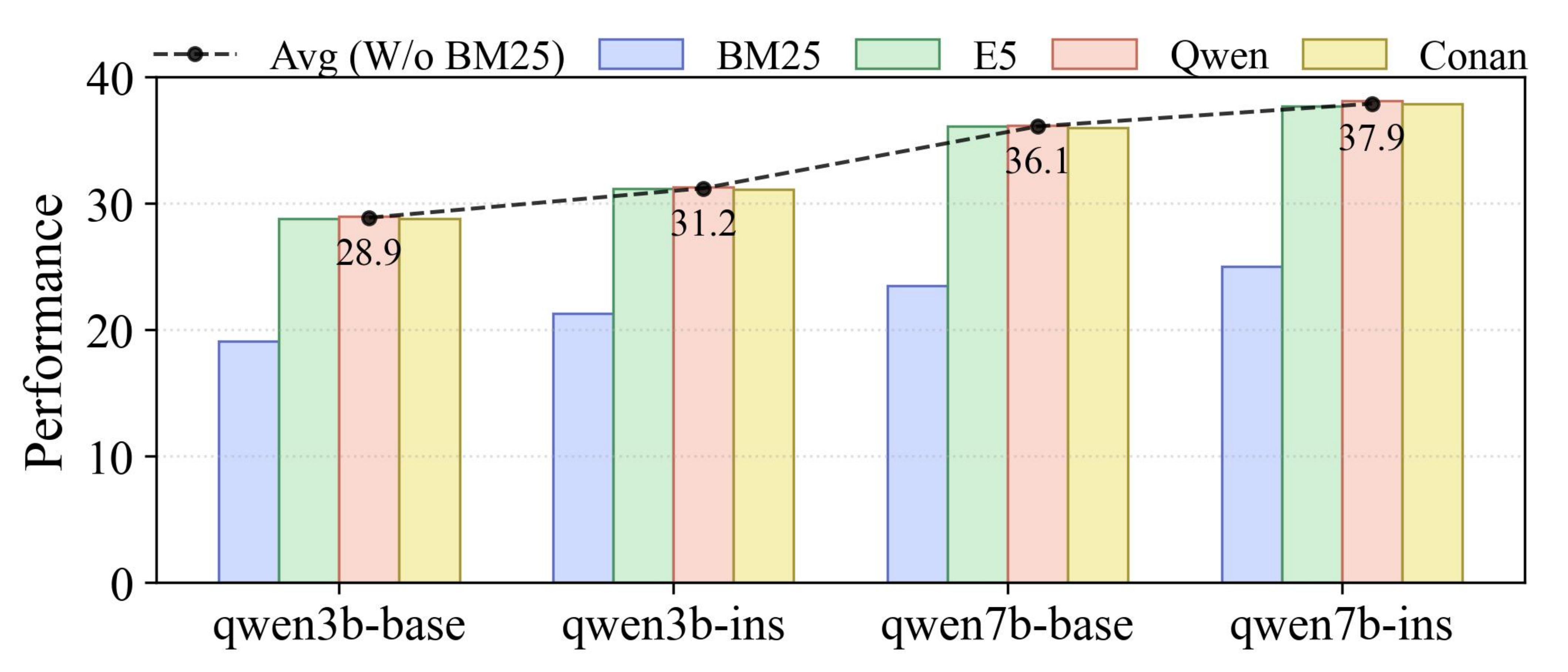}
    \caption{\textbf{Ablation study on embedding and base/instruction models.} We evaluate our method on the Wiki18 corpus across different backbone and embedding models over all datasets. The dashed line denotes the mean performance (excluding BM25).}
    \label{fig:abl_embed}
\end{figure}

\begin{table*}[h]
\caption{\textbf{Ablation study: reranker-only intervention vs.\ RL-trained ReSeek (Qwen2.5-7B).} ``Reranker-only (no RL)'' uses bge-reranker-large scores directly at inference without RL training; ``Prompt-only (no RL)'' uses the same \textsc{judge} prompt as ReSeek but without RL training.}
\label{tab:ablation_prompt}
\centering
\resizebox{\textwidth}{!}{
\begin{tabular}{lccccccccc}
\toprule[1.25pt]
\textbf{Method} & \textbf{NQ} & \textbf{TriviaQA} & \textbf{PopQA} & \textbf{HotpotQA} & \textbf{2Wiki} & \textbf{Musique} & \textbf{Bamboogle} & \textbf{FictionalHot} & \textbf{Avg.} \\
\midrule
Search-R1 (baseline) & 0.393 & 0.610 & 0.397 & 0.370 & 0.414 & 0.146 & 0.281 & 0.034 & 0.342 \\
+ Reranker-only (no RL) & 0.408 & 0.625 & 0.415 & 0.382 & 0.420 & 0.155 & 0.298 & 0.040 & 0.354 \\
+ Prompt-only (no RL) & 0.401 & 0.618 & 0.405 & 0.378 & 0.418 & 0.150 & 0.290 & 0.036 & 0.349 \\
\rowcolor{aliceblue!60}
ReSeek (GRPO, full) & \textbf{0.469} & \textbf{0.640} & \textbf{0.501} & \textbf{0.389} & \textbf{0.382} & \textbf{0.185} & \textbf{0.392} & \textbf{0.061} & \textbf{0.377} \\
\bottomrule[1.25pt]
\end{tabular}}
\end{table*}

\begin{table*}[h!]
    \centering
    \caption{Comparison of ReSeek performance using a static retrieval corpus versus a real-world search engine (Google Search). The `(-real)' suffix denotes experiments with the Google Search API.}
    \label{tab:real-search}
    \vspace{0.5em}
    \setlength{\tabcolsep}{4pt}
    \renewcommand{\arraystretch}{1.2}
    \resizebox{\textwidth}{!}{
    \begin{tabular}{lccccccccc}
        \toprule
        \textbf{Method (ReSeek with)} & \textbf{NQ} & \textbf{TriviaQA} & \textbf{PopQA} & \textbf{HotpotQA} & \textbf{2wiki} & \textbf{Musique} & \textbf{Bamboogle} & \textbf{FictionalHot} & \textbf{Avg.} \\
        \midrule
        Qwen2.5-3B-Instruct & 0.415 & 0.553 & 0.434 & 0.328 & 0.298 & 0.103 & 0.304 & \textbf{0.059} & 0.312 \\
        Qwen2.5-3B-Instruct (-real) & \textbf{0.462} & \textbf{0.605} & \textbf{0.498} & \textbf{0.371} & \textbf{0.345} & \textbf{0.138} & \textbf{0.355} & 0.003 & \textbf{0.354} \\
        \hdashline
        Qwen2.5-7B-Instruct & 0.469 & 0.640 & 0.501 & 0.389 & 0.382 & 0.185 & 0.392 & \textbf{0.061} & 0.377 \\
        Qwen2.5-7B-Instruct (-real) & \textbf{0.511} & \textbf{0.695} & \textbf{0.557} & \textbf{0.442} & \textbf{0.428} & \textbf{0.224} & \textbf{0.460} & 0.003 & \textbf{0.422} \\
        \bottomrule
    \end{tabular}}
\end{table*}

\noindent\myparagraph{Sensitivity to the retrieval encoder.}
We ablate the search embedding on the Wiki18 corpus by training a dedicated agent for each retriever. Throughout this process, we hold the rest of the pipeline fixed and evaluate the results, covering both base and instruction backbones.
As shown in Figure~\ref{fig:abl_embed}, BM25~\cite{robertson2009probabilistic} consistently underperforms the dense retrievers, reflecting lexical mismatch and limited semantic coverage. Among dense encoders, E5~\cite{wang2022text}, Qwen~\cite{zhang2025qwen3}, and Conan~\cite{li2025conan} are close, with Qwen slightly ahead of E5. 
Because our datasets are entity-centric, retrieval isn’t particularly hard and performance changes little with reasonably capable embeddings.

\noindent\myparagraph{Base vs. instruction-tuned backbones.}
We focus on the difference between base and instruction-tuned backbones. 
As shown in Figure~\ref{fig:abl_embed}, averaging over dense embeddings (excluding BM25), instruction-tuned models consistently outperform their base counterparts: qwen3b shows +2.3 points and qwen7b shows +1.8 points. 
This gap arises because instruction-tuned models adhere more faithfully to structured prompting and tool-use conventions, which our method relies on to compose queries, filter evidence, and update intermediate states. 
Base models are less consistent and therefore perform worse. 
For fairness, we avoid cold-start SFT and prompt-engineering, which could increase base-model performance. 

\begin{figure}[h]
\centering
\includegraphics[width=1.0\columnwidth]{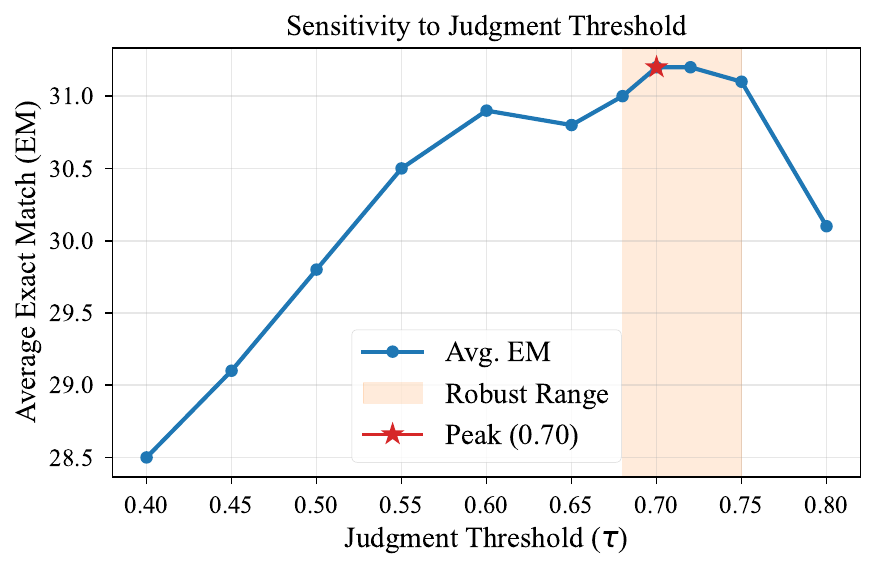}
\caption{\textbf{Sensitivity analysis of judgment threshold $\tau$.} The model performance (Avg. EM) peaks at $\tau=0.70$ and remains robust within the $[0.68, 0.75]$ range.}
\label{fig:threshold_sensitivity}
\end{figure}

\noindent\myparagraph{Sensitivity Analysis of Judgment Threshold}\label{sec:Threshold}
In our framework, the threshold $\tau$ is utilized during the training phase to convert the continuous relevance scores from the external reranker (specifically, `bge-reranker`) into binary judgment labels (``good'' or ``bad'') for supervising the agent. To determine the optimal setting for this hyperparameter, we conducted a sensitivity analysis by varying $\tau$ from 0.40 to 0.80. As shown in Figure \ref{fig:threshold_sensitivity}, the model performance peaks at an average EM of 31.2\% when $\tau=0.70$. Furthermore, the results exhibit stability within the $[0.68, 0.75]$ range, confirming that the training process is robust to minor fluctuations around the selected threshold.

\noindent\myparagraph{Performance with a Real-World Search Engine.}
To evaluate the real-world applicability of ReSeek, we replaced our static retrieval corpus with the Google Search API. As shown in Table~\ref{tab:real-search}, integrating a live search engine provides a substantial performance boost. The average score for the 7B model, for instance, increases from 0.377 to 0.422. This improvement is attributed to the access to fresher and higher-quality information from the web, which is particularly beneficial for knowledge-intensive datasets. Crucially, performance on FictionalHot is not absolute zero due to a small subset of binary (Yes/No) questions. Even without retrieving relevant information (since the facts do not exist on Google), the model has a probability of guessing the correct label by chance. For all specific factoid questions, the accuracy is 0\%, confirming that the dataset is effectively isolated from real-world knowledge.

\begin{figure*}[t]
\begin{center}
\centerline{\includegraphics[width=2.0\columnwidth]{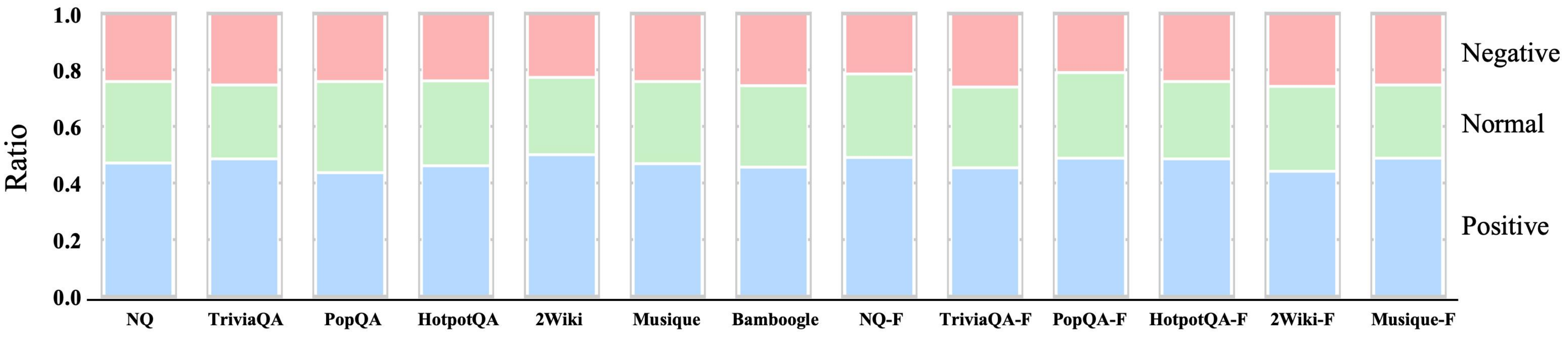}}
\caption{\textbf{Qualitative analysis of our \textsc{JUDGE} action impact.} We categorize each case as `Positive' (beneficial intervention), `Negative' (detrimental intervention), or `Normal'.}
\label{fig:judge_ratio}
\end{center}
\end{figure*}
\subsection{Qualitative Analysis}
To complement our quantitative metrics and provide a deeper understanding beyond aggregate scores, we performed a fine-grained analysis of the judge's decision-making behavior. We established a taxonomy that classifies interventions into three distinct categories: Positive, Negative, and Normal. Specifically, Positive interventions characterize scenarios where the judge actively contributes to success, either by correctly validating relevant evidence or by effectively shielding the agent from misleading context. In contrast, Negative interventions denote failure modes where the judge approves the retrieved content, yet the model subsequently fails to generate the correct response. Finally, the Normal category encompasses neutral instances where the judge's intervention serves as a baseline, neither significantly aiding nor hindering the final outcome.

As visually illustrated in Figure \ref{fig:judge_ratio}, the empirical distribution reveals a promising trend: the proportion of Positive outcomes is substantial across all twelve benchmark settings, consistently comprising 40–50\% of the total cases. Conversely, Negative outcomes are strictly suppressed below 25\%. This significant margin highlights a favorable signal-to-noise ratio, underscoring the mechanism's reliability in guiding correct reasoning paths while effectively minimizing the risk of error propagation.

\section{Limitations}\label{sec:limitations}
ReSeek uses an answer-conditioned signal to supervise the \textsc{judge} action, where the utility of retrieved evidence is estimated by its semantic relevance to the ground-truth answer. This design is effective and efficient for standard QA benchmarks, where answer strings provide a reliable supervision source and useful evidence often co-occurs with the final answer or its supporting documents. In such settings, low evidence-answer similarity naturally encourages the agent to continue searching rather than prematurely commit to incomplete context, and our asymmetric penalty further favors ``search again" over accepting noisy evidence.

However, this reward signal may become weaker in deep-search scenarios where useful intermediate evidence is only indirectly related to the final answer, such as cases requiring long evidence chains or abstract bridge entities. Although ReSeek still improves over Search-R1 on multi-hop datasets, richer supervision could further strengthen its judgment capability in these settings. A promising direction is to leverage per-step gold evidence, when available, to construct evidence-level reward signals. More broadly, the ReSeek framework is compatible with alternative judgment signals, such as LLM-as-a-judge or evidence-conditioned rewards, which could further improve self-correction in complex multi-hop reasoning.

\section{Conclusion}\label{sec:conclusion}

In this paper, we introduced ReSeek, a self-correcting framework that enables search agents to recover from intermediate reasoning errors.
ReSeek empowers agents with a dynamic self-correction mechanism centered on a \textsc{judge} action, allowing them to pause, evaluate information, and adapt their strategy mid-episode.
This process is guided by a dense, instructive reward function that provides fine-grained feedback on both the factual correctness and contextual utility of information.
Furthermore, to ensure rigorous evaluation, we proposed FictionalHot, a contamination-resistant benchmark testing procedural reasoning.
Our extensive experiments demonstrate that ReSeek significantly outperforms SOTA baselines across a wide range of open-domain QA tasks, particularly in complex multi-hop scenarios. 
Qualitative analysis further confirmed that our \textsc{judge} mechanism consistently provides substantial positive benefits while incurring minimal negative side effects, validating its reliability. 
We hope that the principle of in-episode self-correction demonstrated in ReSeek will contribute to the creation of more robust and reflective search agents.

\section{Impact Statement} This paper presents work whose goal is to advance the field of machine learning, specifically by improving the reliability and factual grounding of search-augmented Large Language Models (LLMs). A primary motivation of our framework, \textbf{ReSeek}, is to mitigate the risk of factual hallucinations and error propagation, thereby contributing to the development of more trustworthy and safe AI systems. Regarding data ethics, our introduced benchmark, \textbf{FictionalHot}, is constructed entirely from fictional entities to strictly avoid Personally Identifiable Information (PII) or privacy concerns. While our method aims to enhance factual accuracy, we acknowledge that the underlying LLMs may still inherit and amplify societal biases present in their pre-training corpora. Consequently, we recommend that any real-world deployment of such systems be accompanied by rigorous safety testing and bias auditing.

\bibliography{example_paper}

@article{Joshi_Choi_Weld_Zettlemoyer_2017,  
 title={TriviaQA: A Large Scale Distantly Supervised Challenge Dataset for Reading Comprehension}, 
 journal={Cornell University - arXiv}, 
 author={Joshi, Mandar and Choi, Eunsol and Weld, DanielS. and Zettlemoyer, Luke}, 
 year={2017}, 
 month={May}, 
 language={en-US} 
 }

@article{Kwiatkowski_Palomaki_Redfield_Collins_Parikh_Alberti_Epstein_Polosukhin_Devlin_Lee_e2019,  
 title={Natural Questions: A Benchmark for Question Answering Research}, 
 url={http://dx.doi.org/10.1162/tacl_a_00276}, 
 DOI={10.1162/tacl_a_00276}, 
 journal={Transactions of the Association for Computational Linguistics}, 
 author={Kwiatkowski, Tom and Palomaki, Jennimaria and Redfield, Olivia and Collins, Michael and Parikh, Ankur and Alberti, Chris and Epstein, Danielle and Polosukhin, Illia and Devlin, Jacob and Lee, Kenton and Toutanova, Kristina and Jones, Llion and Kelcey, Matthew and Chang, Ming-Wei and Dai, Andrew M. and Uszkoreit, Jakob and Le, Quoc and Petrov, Slav}, 
 year={2019}, 
 month={Nov}, 
 pages={453–466}, 
 language={en-US} 
 }

@article{lewis2020retrieval,
  title={Retrieval-augmented generation for knowledge-intensive nlp tasks},
  author={Lewis, Patrick and Perez, Ethan and Piktus, Aleksandra and Petroni, Fabio and Karpukhin, Vladimir and Goyal, Naman and K{\"u}ttler, Heinrich and Lewis, Mike and Yih, Wen-tau and Rockt{\"a}schel, Tim and others},
  journal={Advances in neural information processing systems},
  volume={33},
  pages={9459--9474},
  year={2020}
}

@article{brown2020language,
  title={Language models are few-shot learners},
  author={Brown, Tom and Mann, Benjamin and Ryder, Nick and Subbiah, Melanie and Kaplan, Jared D and Dhariwal, Prafulla and Neelakantan, Arvind and Shyam, Pranav and Sastry, Girish and Askell, Amanda and others},
  journal={Advances in neural information processing systems},
  volume={33},
  pages={1877--1901},
  year={2020}
}

@article{ouyang2022training,
  title={Training language models to follow instructions with human feedback},
  author={Ouyang, Long and Wu, Jeffrey and Jiang, Xu and Almeida, Diogo and Wainwright, Carroll and Mishkin, Pamela and Zhang, Chong and Agarwal, Sandhini and Slama, Katarina and Ray, Alex and others},
  journal={Advances in neural information processing systems},
  volume={35},
  pages={27730--27744},
  year={2022}
}

@article{robertson2009probabilistic,
  title={The probabilistic relevance framework: BM25 and beyond},
  author={Robertson, Stephen and Zaragoza, Hugo and others},
  journal={Foundations and Trends{\textregistered} in Information Retrieval},
  volume={3},
  number={4},
  pages={333--389},
  year={2009},
  publisher={Now Publishers, Inc.}
}

@article{zhang2025qwen3,
  title={Qwen3 Embedding: Advancing Text Embedding and Reranking Through Foundation Models},
  author={Zhang, Yanzhao and Li, Mingxin and Long, Dingkun and Zhang, Xin and Lin, Huan and Yang, Baosong and Xie, Pengjun and Yang, An and Liu, Dayiheng and Lin, Junyang and others},
  journal={arXiv preprint arXiv:2506.05176},
  year={2025}
}

@article{li2025conan,
  title={Conan-Embedding-v2: Training an LLM from Scratch for Text Embeddings},
  author={Li, Shiyu and Tang, Yang and Liu, Ruijie and Chen, Shi-Zhe and Chen, Xi},
  journal={arXiv preprint arXiv:2509.12892},
  year={2025}
}

@article{wang2022text,
  title={Text embeddings by weakly-supervised contrastive pre-training},
  author={Wang, Liang and Yang, Nan and Huang, Xiaolong and Jiao, Binxing and Yang, Linjun and Jiang, Daxin and Majumder, Rangan and Wei, Furu},
  journal={arXiv preprint arXiv:2212.03533},
  year={2022}
}

@misc{qwen2025qwen25technicalreport,
      title={Qwen2.5 Technical Report}, 
      author={Qwen and : and An Yang and Baosong Yang and Beichen Zhang and Binyuan Hui and Bo Zheng and Bowen Yu and Chengyuan Li and Dayiheng Liu and Fei Huang and Haoran Wei and Huan Lin and Jian Yang and Jianhong Tu and Jianwei Zhang and Jianxin Yang and Jiaxi Yang and Jingren Zhou and Junyang Lin and Kai Dang and Keming Lu and Keqin Bao and Kexin Yang and Le Yu and Mei Li and Mingfeng Xue and Pei Zhang and Qin Zhu and Rui Men and Runji Lin and Tianhao Li and Tianyi Tang and Tingyu Xia and Xingzhang Ren and Xuancheng Ren and Yang Fan and Yang Su and Yichang Zhang and Yu Wan and Yuqiong Liu and Zeyu Cui and Zhenru Zhang and Zihan Qiu},
      year={2025},
      eprint={2412.15115},
      archivePrefix={arXiv},
      primaryClass={cs.CL},
      url={https://arxiv.org/abs/2412.15115}, 
}

@inproceedings{karpukhin2020dense,
  title={Dense Passage Retrieval for Open-Domain Question Answering.},
  author={Karpukhin, Vladimir and Oguz, Barlas and Min, Sewon and Lewis, Patrick SH and Wu, Ledell and Edunov, Sergey and Chen, Danqi and Yih, Wen-tau},
  booktitle={EMNLP (1)},
  pages={6769--6781},
  year={2020}
}

@article{chen2025learning,
  title={Learning to reason with search for llms via reinforcement learning},
  author={Chen, Mingyang and Li, Tianpeng and Sun, Haoze and Zhou, Yijie and Zhu, Chenzheng and Wang, Haofen and Pan, Jeff Z and Zhang, Wen and Chen, Huajun and Yang, Fan and others},
  journal={arXiv preprint arXiv:2503.19470},
  year={2025}
}

@article{yang2025qwen3,
  title={Qwen3 technical report},
  author={Yang, An and Li, Anfeng and Yang, Baosong and Zhang, Beichen and Hui, Binyuan and Zheng, Bo and Yu, Bowen and Gao, Chang and Huang, Chengen and Lv, Chenxu and others},
  journal={arXiv preprint arXiv:2505.09388},
  year={2025}
}

@article{guo2025deepseek,
  title={Deepseek-r1: Incentivizing reasoning capability in llms via reinforcement learning},
  author={Guo, Daya and Yang, Dejian and Zhang, Haowei and Song, Junxiao and Zhang, Ruoyu and Xu, Runxin and Zhu, Qihao and Ma, Shirong and Wang, Peiyi and Bi, Xiao and others},
  journal={arXiv preprint arXiv:2501.12948},
  year={2025}
}

@article{zhao2024recommender,
  title={Recommender systems in the era of large language models (llms)},
  author={Zhao, Zihuai and Fan, Wenqi and Li, Jiatong and Liu, Yunqing and Mei, Xiaowei and Wang, Yiqi and Wen, Zhen and Wang, Fei and Zhao, Xiangyu and Tang, Jiliang and others},
  journal={IEEE Transactions on Knowledge and Data Engineering},
  volume={36},
  number={11},
  pages={6889--6907},
  year={2024},
  publisher={IEEE}
}

@inproceedings{borgeaud2022improving,
  title={Improving language models by retrieving from trillions of tokens},
  author={Borgeaud, Sebastian and Mensch, Arthur and Hoffmann, Jordan and Cai, Trevor and Rutherford, Eliza and Millican, Katie and Van Den Driessche, George Bm and Lespiau, Jean-Baptiste and Damoc, Bogdan and Clark, Aidan and others},
  booktitle={International conference on machine learning},
  pages={2206--2240},
  year={2022},
  organization={PMLR}
}

@inproceedings{maleki2024ai,
  title={AI hallucinations: a misnomer worth clarifying},
  author={Maleki, Negar and Padmanabhan, Balaji and Dutta, Kaushik},
  booktitle={2024 IEEE conference on artificial intelligence (CAI)},
  pages={133--138},
  year={2024},
  organization={IEEE}
}

@article{li2025webthinker,
  title={Webthinker: Empowering large reasoning models with deep research capability},
  author={Li, Xiaoxi and Jin, Jiajie and Dong, Guanting and Qian, Hongjin and Zhu, Yutao and Wu, Yongkang and Wen, Ji-Rong and Dou, Zhicheng},
  journal={arXiv preprint arXiv:2504.21776},
  year={2025}
}

@article{zheng2025deepresearcher,
  title={Deepresearcher: Scaling deep research via reinforcement learning in real-world environments},
  author={Zheng, Yuxiang and Fu, Dayuan and Hu, Xiangkun and Cai, Xiaojie and Ye, Lyumanshan and Lu, Pengrui and Liu, Pengfei},
  journal={arXiv preprint arXiv:2504.03160},
  year={2025}
}

@article{jin2025search,
  title={Search-r1: Training llms to reason and leverage search engines with reinforcement learning},
  author={Jin, Bowen and Zeng, Hansi and Yue, Zhenrui and Yoon, Jinsung and Arik, Sercan and Wang, Dong and Zamani, Hamed and Han, Jiawei},
  journal={arXiv preprint arXiv:2503.09516},
  year={2025}
}

@article{luo2025large,
  title={Large language model agent: A survey on methodology, applications and challenges},
  author={Luo, Junyu and Zhang, Weizhi and Yuan, Ye and Zhao, Yusheng and Yang, Junwei and Gu, Yiyang and Wu, Bohan and Chen, Binqi and Qiao, Ziyue and Long, Qingqing and others},
  journal={arXiv preprint arXiv:2503.21460},
  year={2025}
}

@article{chatgpt,
  title={Introducing chatgpt},
  author={OpenAI},
  journal={CoRR},
  url={https://openai.com/blog/chatgpt},
  year={2022},
}

@article{zhang2025llm,
  title={Llm hallucinations in practical code generation: Phenomena, mechanism, and mitigation},
  author={Zhang, Ziyao and Wang, Chong and Wang, Yanlin and Shi, Ensheng and Ma, Yuchi and Zhong, Wanjun and Chen, Jiachi and Mao, Mingzhi and Zheng, Zibin},
  journal={Proceedings of the ACM on Software Engineering},
  volume={2},
  number={ISSTA},
  pages={481--503},
  year={2025},
  publisher={ACM New York, NY, USA}
}

@article{songR1SearcherIncentivizingSearch2025,
  title = {R1-{{Searcher}}: {{Incentivizing}} the {{Search Capability}} in {{LLMs}} via {{Reinforcement Learning}}},
  shorttitle = {R1-{{Searcher}}},
  author = {Song, Huatong and Jiang, Jinhao and Min, Yingqian and Chen, Jie and Chen, Zhipeng and Zhao, Wayne Xin and Fang, Lei and Wen, Ji-Rong},
  date = {2025-03-18},
  eprint = {2503.05592},
  eprinttype = {arXiv},
  eprintclass = {cs},
  doi = {10.48550/arXiv.2503.05592},
  url = {http://arxiv.org/abs/2503.05592},
  urldate = {2025-07-09},
  pubstate = {prepublished}
}

@article{songR1SearcherIncentivizingDynamic2025,
  title = {R1-{{Searcher}}++: {{Incentivizing}} the {{Dynamic Knowledge Acquisition}} of {{LLMs}} via {{Reinforcement Learning}}},
  shorttitle = {R1-{{Searcher}}++},
  author = {Song, Huatong and Jiang, Jinhao and Tian, Wenqing and Chen, Zhipeng and Wu, Yuhuan and Zhao, Jiahao and Min, Yingqian and Zhao, Wayne Xin and Fang, Lei and Wen, Ji-Rong},
  date = {2025-05-22},
  eprint = {2505.17005},
  eprinttype = {arXiv},
  eprintclass = {cs},
  doi = {10.48550/arXiv.2505.17005},
  url = {http://arxiv.org/abs/2505.17005},
  urldate = {2025-07-28},
  pubstate = {prepublished}
}

@article{sunZeroSearchIncentivizeSearch2025,
  title = {{{ZeroSearch}}: {{Incentivize}} the {{Search Capability}} of {{LLMs}} without {{Searching}}},
  shorttitle = {{{ZeroSearch}}},
  author = {Sun, Hao and Qiao, Zile and Guo, Jiayan and Fan, Xuanbo and Hou, Yingyan and Jiang, Yong and Xie, Pengjun and Zhang, Yan and Huang, Fei and Zhou, Jingren},
  date = {2025-05-16},
  eprint = {2505.04588},
  eprinttype = {arXiv},
  eprintclass = {cs},
  doi = {10.48550/arXiv.2505.04588},
  url = {http://arxiv.org/abs/2505.04588},
  urldate = {2025-07-04},
  pubstate = {prepublished}
}

@inproceedings{hoConstructingMultihopQA2020,
  title = {Constructing {{A Multi-hop QA Dataset}} for {{Comprehensive Evaluation}} of {{Reasoning Steps}}},
  booktitle = {Proceedings of the 28th {{International Conference}} on {{Computational Linguistics}}},
  author = {Ho, Xanh and Duong Nguyen, Anh-Khoa and Sugawara, Saku and Aizawa, Akiko},
  date = {2020},
  pages = {6609--6625},
  publisher = {International Committee on Computational Linguistics},
  location = {Barcelona, Spain (Online)},
  doi = {10.18653/v1/2020.coling-main.580},
  url = {https://www.aclweb.org/anthology/2020.coling-main.580},
  urldate = {2025-09-22},
  eventtitle = {Proceedings of the 28th {{International Conference}} on {{Computational Linguistics}}}
}

@online{mallenWhenNotTrust2022,
  title = {When {{Not}} to {{Trust Language Models}}: {{Investigating Effectiveness}} of {{Parametric}} and {{Non-Parametric Memories}}},
  shorttitle = {When {{Not}} to {{Trust Language Models}}},
  author = {Mallen, Alex and Asai, Akari and Zhong, Victor and Das, Rajarshi and Khashabi, Daniel and Hajishirzi, Hannaneh},
  date = {2022},
  doi = {10.48550/ARXIV.2212.10511},
  url = {https://arxiv.org/abs/2212.10511},
  urldate = {2025-09-22},
  pubstate = {prepublished},
  version = {4}
}

@online{pressMeasuringNarrowingCompositionality2022,
  title = {Measuring and {{Narrowing}} the {{Compositionality Gap}} in {{Language Models}}},
  author = {Press, Ofir and Zhang, Muru and Min, Sewon and Schmidt, Ludwig and Smith, Noah A. and Lewis, Mike},
  date = {2022},
  doi = {10.48550/ARXIV.2210.03350},
  url = {https://arxiv.org/abs/2210.03350},
  urldate = {2025-09-22},
  pubstate = {prepublished},
  version = {3}
}

@article{trivediMuSiQueMultihopQuestions2022,
  title = {{{MuSiQue}}: {{Multihop Questions}} via {{Single-hop Question Composition}}},
  shorttitle = {{{MuSiQue}}},
  author = {Trivedi, Harsh and Balasubramanian, Niranjan and Khot, Tushar and Sabharwal, Ashish},
  date = {2022-05-04},
  journaltitle = {Transactions of the Association for Computational Linguistics},
  volume = {10},
  pages = {539--554},
  issn = {2307-387X},
  doi = {10.1162/tacl_a_00475},
  urldate = {2025-09-22}
}

@inproceedings{2025mdbench,
  title={Mdbench: A synthetic multi-document reasoning benchmark generated with knowledge guidance},
  author={Peper, Joseph J and Qiu, Wenzhao and Payani, Ali and Wang, Lu},
  booktitle={Findings of the Association for Computational Linguistics: ACL 2025},
  pages={25592--25621},
  year={2025}
}

@article{wu2025hiprag,
  title={Hiprag: hierarchical process rewards for efficient agentic retrieval augmented generation},
  author={Wu, Peilin and Zhang, Mian and Wan, Kun and Zhao, Wentian and He, Kaiyu and Du, Xinya and Chen, Zhiyu},
  journal={arXiv preprint arXiv:2510.07794},
  year={2025}
}

@inproceedings{xi2026agentprm,
  title={Agentprm: Process reward models for llm agents via step-wise promise and progress},
  author={Xi, Zhiheng and Liao, Chenyang and Li, Guanyu and Zhang, Zhihao and Chen, Wenxiang and Wang, Binghai and Jin, Senjie and Zhou, Yuhao and Guan, Jian and Wu, Wei and others},
  booktitle={Proceedings of the ACM Web Conference 2026},
  pages={4184--4195},
  year={2026}
}

@inproceedings{ma2025s2r,
  title={S2r: Teaching llms to self-verify and self-correct via reinforcement learning},
  author={Ma, Ruotian and Wang, Peisong and Liu, Cheng and Liu, Xingyan and Chen, Jiaqi and Zhang, Bang and Zhou, Xin and Du, Nan and Li, Jia},
  booktitle={Proceedings of the 63rd Annual Meeting of the Association for Computational Linguistics (Volume 1: Long Papers)},
  pages={22632--22654},
  year={2025}
}

@inproceedings{yangHotpotQADatasetDiverse2018,
  title = {{{HotpotQA}}: {{A Dataset}} for {{Diverse}}, {{Explainable Multi-hop Question Answering}}},
  shorttitle = {{{HotpotQA}}},
  booktitle = {Proceedings of the 2018 {{Conference}} on {{Empirical Methods}} in {{Natural Language Processing}}},
  author = {Yang, Zhilin and Qi, Peng and Zhang, Saizheng and Bengio, Yoshua and Cohen, William and Salakhutdinov, Ruslan and Manning, Christopher D.},
  date = {2018},
  pages = {2369--2380},
  publisher = {Association for Computational Linguistics},
  location = {Brussels, Belgium},
  doi = {10.18653/v1/D18-1259},
  url = {http://aclweb.org/anthology/D18-1259},
  urldate = {2025-09-22},
  eventtitle = {Proceedings of the 2018 {{Conference}} on {{Empirical Methods}} in {{Natural Language Processing}}}
}

@article{Zhang2023SirensSI,
  title={Siren's Song in the AI Ocean: A Survey on Hallucination in Large Language Models},
  author={Yue Zhang and Yafu Li and Leyang Cui and Deng Cai and Lemao Liu and Tingchen Fu and Xinting Huang and Enbo Zhao and Yu Zhang and Yulong Chen and Longyue Wang and Anh Tuan Luu and Wei Bi and Freda Shi and Shuming Shi},
  journal={ArXiv},
  year={2023},
  volume={abs/2309.01219},
  url={https://api.semanticscholar.org/CorpusID:261530162}
}

@misc{qin2024toolrl,
      title={ToolRL: Reward is All Tool Learning Needs}, 
      author={Cheng Qian and Emre Can Acikgoz and Qi He and Hongru Wang and Xiusi Chen and Dilek Hakkani-Tür and Gokhan Tur and Heng Ji},
      year={2025},
      eprint={2504.13958},
      archivePrefix={arXiv},
      primaryClass={cs.LG},
      url={https://arxiv.org/abs/2504.13958}, 
}

@inproceedings{ICLR2025_80790082,
 author = {Feng, Huawen and ZekunYao and Zheng, Junhao and Ma, Qianli},
 booktitle = {International Conference on Representation Learning},
 editor = {Y. Yue and A. Garg and N. Peng and F. Sha and R. Yu},
 pages = {51866--51884},
 title = {Training Large Language Models for Retrieval-Augmented Question Answering through Backtracking Correction},
 volume = {2025},
 year = {2025}
}

@article{ji2023survey,
   title={Survey of Hallucination in Natural Language Generation},
   volume={55},
   ISSN={1557-7341},
   url={http://dx.doi.org/10.1145/3571730},
   DOI={10.1145/3571730},
   number={12},
   journal={ACM Computing Surveys},
   publisher={Association for Computing Machinery (ACM)},
   author={Ji, Ziwei and Lee, Nayeon and Frieske, Rita and Yu, Tiezheng and Su, Dan and Xu, Yan and Ishii, Etsuko and Bang, Ye Jin and Madotto, Andrea and Fung, Pascale},
   year={2023},
   month=mar, pages={1–38} }

@article{bang2023multitask,
  title={A Multitask, Multilingual, Multimodal Evaluation of ChatGPT on Reasoning, Hallucination, and Interactivity},
  author={Yejin Bang and Samuel Cahyawijaya and Nayeon Lee and Wenliang Dai and Dan Su and Bryan Wilie and Holy Lovenia and Ziwei Ji and Tiezheng Yu and Willy Chung and Quyet V. Do and Yan Xu and Pascale Fung},
  journal={ArXiv},
  year={2023},
  volume={abs/2302.04023},
  url={https://api.semanticscholar.org/CorpusID:256662612}
}

@inproceedings{guu2020realm,
author = {Guu, Kelvin and Lee, Kenton and Tung, Zora and Pasupat, Panupong and Chang, Ming-Wei},
title = {REALM: retrieval-augmented language model pre-training},
year = {2020},
publisher = {JMLR.org},
booktitle = {Proceedings of the 37th International Conference on Machine Learning},
articleno = {368},
numpages = {10},
series = {ICML'20}
}

@article{yuan2025s,
  title={What's Behind PPO's Collapse in Long-CoT? Value Optimization Holds the Secret},
  author={Yuan, Yufeng and Yue, Yu and Zhu, Ruofei and Fan, Tiantian and Yan, Lin},
  journal={arXiv preprint arXiv:2503.01491},
  year={2025}
}

@misc{bge_embedding,
      title={C-Pack: Packaged Resources To Advance General Chinese Embedding}, 
      author={Shitao Xiao and Zheng Liu and Peitian Zhang and Niklas Muennighoff},
      year={2023},
      eprint={2309.07597},
      archivePrefix={arXiv},
      primaryClass={cs.CL}
}

@inproceedings{FlashRAG,
  author       = {Jiajie Jin and
                  Yutao Zhu and
                  Zhicheng Dou and
                  Guanting Dong and
                  Xinyu Yang and
                  Chenghao Zhang and
                  Tong Zhao and
                  Zhao Yang and
                  Ji{-}Rong Wen},
  editor       = {Guodong Long and
                  Michale Blumestein and
                  Yi Chang and
                  Liane Lewin{-}Eytan and
                  Zi Helen Huang and
                  Elad Yom{-}Tov},
  title        = {FlashRAG: {A} Modular Toolkit for Efficient Retrieval-Augmented Generation
                  Research},
  booktitle    = {Companion Proceedings of the {ACM} on Web Conference 2025, {WWW} 2025,
                  Sydney, NSW, Australia, 28 April 2025 - 2 May 2025},
  pages        = {737--740},
  publisher    = {{ACM}},
  year         = {2025},
  url          = {https://doi.org/10.1145/3701716.3715313},
  doi          = {10.1145/3701716.3715313}
}

@article{jiang2023active,
  title={Active Retrieval Augmented Generation},
  author={Zhengbao Jiang and Frank F. Xu and Luyu Gao and Zhiqing Sun and Qian Liu and Jane Dwivedi-Yu and Yiming Yang and Jamie Callan and Graham Neubig},
  journal={ArXiv},
  year={2023},
  volume={abs/2305.06983},
  url={https://api.semanticscholar.org/CorpusID:258615731}
}

@article{asai2023self,
  title={Self-RAG: Learning to Retrieve, Generate, and Critique through Self-Reflection},
  author={Akari Asai and Zeqiu Wu and Yizhong Wang and Avirup Sil and Hannaneh Hajishirzi},
  journal={ArXiv},
  year={2023},
  volume={abs/2310.11511},
  url={https://api.semanticscholar.org/CorpusID:264288947}
}

@inproceedings{shinn2023reflexion,
  title={Reflexion: language agents with verbal reinforcement learning},
  author={Noah Shinn and Federico Cassano and Beck Labash and Ashwin Gopinath and Karthik Narasimhan and Shunyu Yao},
  booktitle={Neural Information Processing Systems},
  year={2023},
  url={https://api.semanticscholar.org/CorpusID:258833055}
}

@article{gao2024zerosearch,
  title={ZeroSearch: Incentivize the Search Capability of LLMs without Searching},
  author={Hao Sun and Zile Qiao and Jiayan Guo and Xuanbo Fan and Yingyan Hou and Yong Jiang and Pengjun Xie and Yan Zhang and Fei Huang and Jingren Zhou},
  journal={ArXiv},
  year={2025},
  volume={abs/2505.04588},
  url={https://api.semanticscholar.org/CorpusID:278367823}
}

@article{chung2024scaling,
  title={Scaling instruction-finetuned language models},
  author={Chung, Hyung Won and Hou, Le and Longpre, Shayne and Zoph, Barret and Tay, Yi and Fedus, William and Li, Yunxuan and Wang, Xuezhi and Dehghani, Mostafa and Brahma, Siddhartha and others},
  journal={Journal of Machine Learning Research},
  volume={25},
  number={70},
  pages={1--53},
  year={2024}
}

@article{luo2024search,
  title={Search-o1: Agentic Search-Enhanced Large Reasoning Models},
  author={Xiaoxi Li and Guanting Dong and Jiajie Jin and Yuyao Zhang and Yujia Zhou and Yutao Zhu and Peitian Zhang and Zhicheng Dou},
  journal={ArXiv},
  year={2025},
  volume={abs/2501.05366},
  url={https://api.semanticscholar.org/CorpusID:275405676}
}

@article{yao2022react,
  title={ReAct: Synergizing Reasoning and Acting in Language Models},
  author={Yao, Shunyu and Zhao, Jeffrey and Yu, Dian and Du, Nan and Shafran, Izhak and Narasimhan, Karthik and Cao, Yuan},
  journal={arXiv preprint arXiv:2210.03629},
  year={2022}
}

@misc{deng2023agent,
      title={Agent RL Scaling Law: Agent RL with Spontaneous Code Execution for Mathematical Problem Solving}, 
      author={Xinji Mai and Haotian Xu and Zhong-Zhi Li and Xing W and Weinong Wang and Jian Hu and Yingying Zhang and Wenqiang Zhang},
      year={2025},
      eprint={2505.07773},
      archivePrefix={arXiv},
      primaryClass={cs.AI},
      url={https://arxiv.org/abs/2505.07773}, 
}

@misc{liu2023agent,
      title={AgentTuning: Enabling Generalized Agent Abilities for LLMs}, 
      author={Aohan Zeng and Mingdao Liu and Rui Lu and Bowen Wang and Xiao Liu and Yuxiao Dong and Jie Tang},
      year={2023},
      eprint={2310.12823},
      archivePrefix={arXiv},
      primaryClass={cs.CL},
      url={https://arxiv.org/abs/2310.12823}, 
}

@misc{chen2023fireact,
      title={FireAct: Toward Language Agent Fine-tuning}, 
      author={Baian Chen and Chang Shu and Ehsan Shareghi and Nigel Collier and Karthik Narasimhan and Shunyu Yao},
      year={2023},
      eprint={2310.05915},
      archivePrefix={arXiv},
      primaryClass={cs.CL},
      url={https://arxiv.org/abs/2310.05915}, 
}
\bibliographystyle{icml2026}

\newpage
\appendix
\onecolumn
\section{Appendix}

\subsection{Implementation Details}\label{appendix_details}
We provide a detailed description of our implementation to ensure the reproducibility of our results. Our experiments are built upon the internal \texttt{verl} reinforcement learning framework and executed on a cluster equipped with H20.
.
\myparagraph{Model and Data}
The core of our agent is the \texttt{Qwen2.5-3B-Instruct} model, which serves as a shared backbone for both the policy and value networks. To manage memory consumption during training, we enable gradient checkpointing. The agent was trained on the \texttt{hot\_benchmark} dataset, which is formatted to match the structure of Natural Questions (NQ). For data processing, we set the maximum prompt length to 2048 tokens, the maximum response length for generation to 500 tokens, and the maximum observation length from the environment to 500 tokens.

Our model is fine-tuned  on a unified training set merging the training splits of Natural Questions (NQ, 79,168 samples)  and HotpotQA (90,447 samples), sourced from the FlashRAG dataset.

\myparagraph{Training Algorithms}
In our experiments, we compared two policy optimization algorithms: the standard Proximal Policy Optimization (PPO) and Group Relative Policy Optimization (GRPO). To ensure a fair comparison, both algorithms were trained under an identical hyperparameter configuration. The models were trained for a single epoch. The optimizer was configured with a learning rate of \texttt{1e-5} and a learning rate warmup ratio of 0.285. For policy updates, we used a training batch size of 512 episodes. This batch was processed using PPO mini-batches of size 256, which were further divided into micro-batches of size 64. To stabilize training and prevent the policy from deviating excessively from the reference model, we incorporated a KL divergence penalty with a coefficient ($\beta$) of \texttt{0.001}, calculated using the \texttt{low\_var\_kl} formulation. For credit assignment, we used a discount factor ($\gamma$) of 0.99 and Generalized Advantage Estimation (GAE) with a $\lambda$ of 0.95. During the rollout phase, a temperature of 1.0 was used for action sampling.

\myparagraph{System and Environment}
Our implementation relies on PyTorch and utilizes the \texttt{vllm} library for efficient inference during rollouts. We employed Fully Sharded Data Parallelism (FSDP) with parameter offloading to effectively distribute the model across multiple H20. The experimental environment was configured with a maximum of 4 turns per episode (\texttt{max\_turns=4}). The agent interacts with an external retriever service via an HTTP API, which returns the top 3 (\texttt{topk=3}) most relevant documents for a given query.

\myparagraph{Evaluation Dataset Statistics}
We evaluate on a total of 51,713 samples. The breakdown is as follows: PopQA (14,267), TriviaQA (11,313), NQ (3,610), 2WikiMultiHopQA (12,576), HotpotQA (7,405), Musique (2,417), and Bamboogle (125).

\myparagraph{Baseline Configurations}
We compare against four families of methods:
\begin{itemize}
    \item \textbf{Vanilla Prompting}: Includes zero-shot direct answering and Chain-of-Thought (CoT) reasoning without external tools.
    \item \textbf{Single-pass RAG}: Retrieves top-$k$ passages once and generates the answer conditioned on them.
    \item \textbf{Agentic Search}: Uses ReAct-style planning loops to search and reason iteratively, but without reinforcement learning tuning.
    \item \textbf{RL-tuned Policies}: Includes recent methods like Search-R1, ReSearch, and ZeroSearch that optimize search policies.
\end{itemize}

\myparagraph{Metric Details}
The Exact Match (EM) metric considers a prediction correct if its normalized string matches any reference answer. Normalization involves lowercasing, removing punctuation and articles, and collapsing whitespace.

\subsection{FictionalHot Details}\label{appendix_fictionalhot}

\subsubsection{Data Composition}
FictionalHot was constructed by randomly sampling approximately 10\% of the seed questions from six major benchmarks (excluding Bamboogle due to insufficient sample size). The distribution of source data is detailed in Table~\ref{tab:fictional_composition}.

\begin{table}[h]
\centering
\caption{Composition of the FictionalHot Dataset.}
\label{tab:fictional_composition}
\begin{tabular}{lrr}
\toprule
\textbf{Source Dataset} & \textbf{Original Size} & \textbf{FictionalHot Size} \\
\midrule
PopQA & 14,267 & 1,415 \\
2WikiMultiHopQA & 12,576 & 1,247 \\
TriviaQA & 11,313 & 1,122 \\
HotpotQA & 7,405 & 735 \\
Natural Questions (NQ) & 3,610 & 359 \\
Musique & 2,417 & 238 \\
\midrule
\textbf{Total} & \textbf{51,588} & \textbf{5,116} \\
\bottomrule
\end{tabular}
\end{table}

\subsubsection{Ensuring Consistency and Novelty}
We employed specific design strategies to ensure the synthetic data is both logically sound and factually isolated from real-world knowledge.

\myparagraph{Internal Consistency via Logic Mapping.}
To avoid the hallucinations common in scratch-generated reasoning, we adopted a \textit{logic mapping strategy}. We mapped valid, human-verified reasoning chains from the source datasets directly to the fictional domain.
\begin{itemize}
    \item \textbf{Original Logic:} Entity $A$ (Real) $\xrightarrow{Relation R}$ Entity $B$ (Real).
    \item \textbf{Fictional Logic:} Entity $A'$ (Fictional) $\xrightarrow{Relation R}$ Entity $B'$ (Fictional).
\end{itemize}
Since the underlying causal structure is inherited from high-quality seed datasets, the internal logic remains coherent and solvable by definition.

\myparagraph{Factual Fictionality.}
To prevent conflicts with the model's parametric knowledge, we enforced \textit{Factual Fictionality}. We define a "conflict" as a factual overlap rather than a mere name overlap. Even if a generated name (e.g., "Felix Turner") coincidentally matches a real-world figure, the specific \textbf{(Entity, Event)} tuple is guaranteed to be unique. For instance, while a real person named Felix Turner may exist, he has never won the fictional "Sky Slam Dunk Contest." This ensures the ground truth exists \textit{only} in our synthetic documents.

\subsubsection{Human Evaluation}
To empirically validate the quality of FictionalHot, we conducted a rigorous human evaluation on a random sample of \textbf{500 instances} (approx. 10\% of the dataset). Five annotators with NLP research backgrounds independently reviewed 100 instances each based on a strict verification rubric. The results are summarized in Table~\ref{tab:human_eval}.

The high pass rates validate the robustness of our generation pipeline. Specifically, the few cases flagged during the Novelty check (0.4\%) involved entity names resembling real-world figures, yet the generated documents described completely different events, ensuring no actual factual contradiction. Regarding Consistency, the rare failures (0.8\%) stemmed from minor information ambiguity; however, we note that such ambiguity is also prevalent in real-world corpora (e.g., Wikipedia), rendering these instances realistic retrieval challenges rather than critical errors.
\begin{table*}[h]
\centering
\caption{Human Evaluation Results for FictionalHot Quality.}
\label{tab:human_eval}
\begin{tabular}{lp{8cm}c}
\toprule
\textbf{Metric} & \textbf{Verification Method} & \textbf{Pass Rate} \\
\midrule
\textbf{Factual Novelty} & \textbf{The "Wikipedia Entity Check":} Annotators searched for the fictional entity names in the question directly on Wikipedia. The instance passed if the entity name did \textbf{not} exist as a Wikipedia entry, confirming the entity occupies a unique namespace. & 99.6\% \\
\midrule
\textbf{Logical Consistency} & \textbf{The "Solvability Check":} Annotators read the generated synthetic document. The instance passed if the document provided sufficient and unambiguous evidence to derive the ground truth answer. & 99.2\% \\
\bottomrule
\end{tabular}
\end{table*}

\subsubsection{Consistency Audit}
To further validate FictionalHot's internal consistency beyond the human evaluation in Table~\ref{tab:human_eval}, we conducted an additional 500-sample audit. Among the 500 audited instances, \textbf{484} (96.8\%) were fully consistent, \textbf{8} (1.6\%) had minor issues (e.g., ambiguous but not contradictory phrasing), \textbf{4} (0.8\%) had major contradictions, and \textbf{4} (0.8\%) were ambiguous. This is consistent with the 99.2\% internal-consistency rate reported in Table~\ref{tab:human_eval} and confirms that serious contradictions are rare, though not impossible. Automated consistency checks remain a promising direction for further improvement.

\subsection{PPO vs. GRPO and Base vs. Instruct}\label{appendix_ppo}
\begin{table*}[h!]
    \centering
    \caption{Performance comparison of ReSeek trained with GRPO versus PPO on both base and instruction-tuned models. GRPO consistently outperforms PPO across most datasets and model configurations. The `Avg.` column is the average score across all eight datasets.}
    \label{tab:ppo-grpo}
    \vspace{.6em}
    \setlength{\tabcolsep}{4pt}
    \renewcommand{\arraystretch}{1.2}
    \resizebox{\textwidth}{!}{
    \begin{tabular}{lccccccccc}
        \toprule
        \textbf{Method} & \textbf{NQ} & \textbf{TriviaQA} & \textbf{PopQA} & \textbf{HotpotQA} & \textbf{2wiki} & \textbf{Musique} & \textbf{Bamboogle} & \textbf{FictionalHot} & \textbf{Avg.} \\
        \midrule
        \multicolumn{10}{l}{\textbf{Qwen2.5-7B-Base/Instruct}} \\
        ReSeek-base (GRPO) & 0.4654 & 0.60 & 0.4917 & 0.358 & 0.345 & 0.140 & 0.371 & 0.052 & 0.353 \\
        ReSeek-instruct (GRPO) & \textbf{0.469} & \textbf{0.640} & \textbf{0.501} & \textbf{0.389} & \textbf{0.382} & \textbf{0.185} & \textbf{0.392} &  \textbf{0.061} & \textbf{0.377} \\
        \hdashline
        ReSeek-base (PPO) & 0.391 & 0.565 & 0.418 & 0.320 & 0.317 & 0.112 & 0.345 & 0.03 & 0.299 \\
        ReSeek-instruct (PPO) & 0.432 & 0.610 & 0.473 & 0.365 & 0.358 & 0.159 & 0.366 & 0.055 & 0.352 \\

        \midrule
        \multicolumn{10}{l}{\textbf{Qwen2.5-3B-Base/Instruct}} \\
        ReSeek-base (GRPO) & \textbf{0.421} & \textbf{0.560} & 0.425 & 0.273 & 0.275 & 0.081 & 0.280 & \textbf{0.050} & 0.296 \\
        ReSeek-instruct (GRPO) & 0.415 & \textbf{0.553} & \textbf{0.434} & \textbf{0.328} & \textbf{0.298} & \textbf{0.103} & \textbf{0.304} & 0.059 & \textbf{0.312} \\
        \hdashline
        ReSeek-base (PPO) & 0.362 & 0.495 & 0.381 & 0.275 & 0.251 & 0.065 & 0.255 & 0.03 & 0.297 \\
        ReSeek-instruct (PPO) & 0.385 & 0.525 & 0.410 & 0.301 & 0.277 & 0.088 & 0.281 & 0.052 & 0.290 \\
        \bottomrule
    \end{tabular}}
\end{table*}

\myparagraph{Base vs. Instruction-Tuned Backbones.}
As discussed in the main body, the choice of backbone model significantly impacts performance. The results in Table~\ref{tab:ppo-grpo} confirm that instruction-tuned models generally outperform their base counterparts on most standard QA datasets. This advantage arises because our method relies on the model's ability to interpret structured prompts for composing queries, filtering evidence, and updating states. Instruction-tuned models, having been trained to follow complex instructions, adhere to these conventions more faithfully. In contrast, base models exhibit less consistency in following the structured format, leading to degraded performance.

\myparagraph{Superiority of GRPO over PPO.}
The empirical results also reveal a clear and consistent advantage of GRPO over PPO. This performance gap is not merely incremental but is rooted in a fundamental training stability issue we encountered with PPO. Specifically, when training with PPO, the agent frequently suffered from \textbf{policy collapse}~\cite{yuan2025s}, a known challenge in reinforcement learning, especially for tasks involving long generation horizons.

This issue was particularly acute in our framework due to the long and complex Chain-of-Thought (CoT) reasoning paths. We observed that after an initial learning phase, the PPO policy would abruptly degrade, characterized by a simultaneous and rapid drop in both the reward signal and the policy's entropy. This collapse rendered the model unable to perform the task, as it began generating repetitive or nonsensical outputs. In contrast, GRPO demonstrated significantly greater training stability, successfully navigating the long CoT trajectories without collapsing and achieving steady performance gains. This inherent robustness makes GRPO a far more suitable and reliable algorithm for our complex reasoning task, explaining its superior final performance.

\subsection{More Results on Bigger LLMs}
\label{appendix_bigllm}

To evaluate the scalability and effectiveness of ReSeek on larger and more capable language models, we extend our experiments to include Qwen3-8B and Qwen3-30B-A3B-Thinking-2507. Table~\ref{tab:bigger-llms} presents a comparative analysis of ReSeek applied to a range of models, from 3B to 32B parameters.

\begin{table*}[h!]
    \centering
    \caption{Performance of ReSeek on various instruction-tuned models of increasing scale.}
    \label{tab:bigger-llms}
    \resizebox{\textwidth}{!}{
    \begin{tabular}{lccccccccc}
        \toprule
        \textbf{Method (ReSeek with)} & \textbf{NQ} & \textbf{TriviaQA} & \textbf{PopQA} & \textbf{HotpotQA} & \textbf{2wiki} & \textbf{Musique} & \textbf{Bamboogle} & \textbf{FictionalHot} & \textbf{Avg.} \\
        \midrule
        Qwen2.5-3B-Instruct & 0.415 & 0.553 & 0.434 & 0.328 & 0.298 & 0.103 & 0.304 & 0.059 & 0.312 \\
        Qwen2.5-7B-Instruct & 0.469 & 0.640 & 0.501 & 0.389 & 0.382 & 0.185 & 0.392 & 0.061 & 0.377 \\
        Qwen3-8B & 0.475 & 0.635 & 0.495 & 0.401 & 0.379 & 0.192 & 0.410 & 0.058 & 0.381 \\
        Qwen3-30B-A3B-Thinking-2507 & \textbf{0.495} & \textbf{0.671} & \textbf{0.521} & \textbf{0.455} & \textbf{0.458} & \textbf{0.235} & \textbf{0.560} & \textbf{0.071} & \textbf{0.433} \\
        \bottomrule
    \end{tabular}}
\end{table*}

The results clearly demonstrate the strong scalability of our ReSeek framework. As the model size increases, the overall performance consistently improves. We observe a significant performance leap from the 3B model (0.312 avg.) to the 7B/8B models (~0.380 avg.), and another substantial gain with the 30B model (0.479 avg.).

Notably, the performance between Qwen2.5-7B-Instruct an Qwen3-8B is highly competitive and neck-and-neck, with each model excelling on different datasets. For instance, Qwen2.5-7B-Instruct shows a slight edge on PopQA and 2WikiMQA, while Qwen3-8B performs better on NQ and HotpotQA. This indicates that ReSeek can effectively leverage the distinct strengths of different backbone models. The Qwen3-30B-A3B-Thinking-2507 achieves the best results across almost all datasets, establishing a new level of performance.

\begin{figure*}[h]
    \centering
    \includegraphics[width=0.7\linewidth]{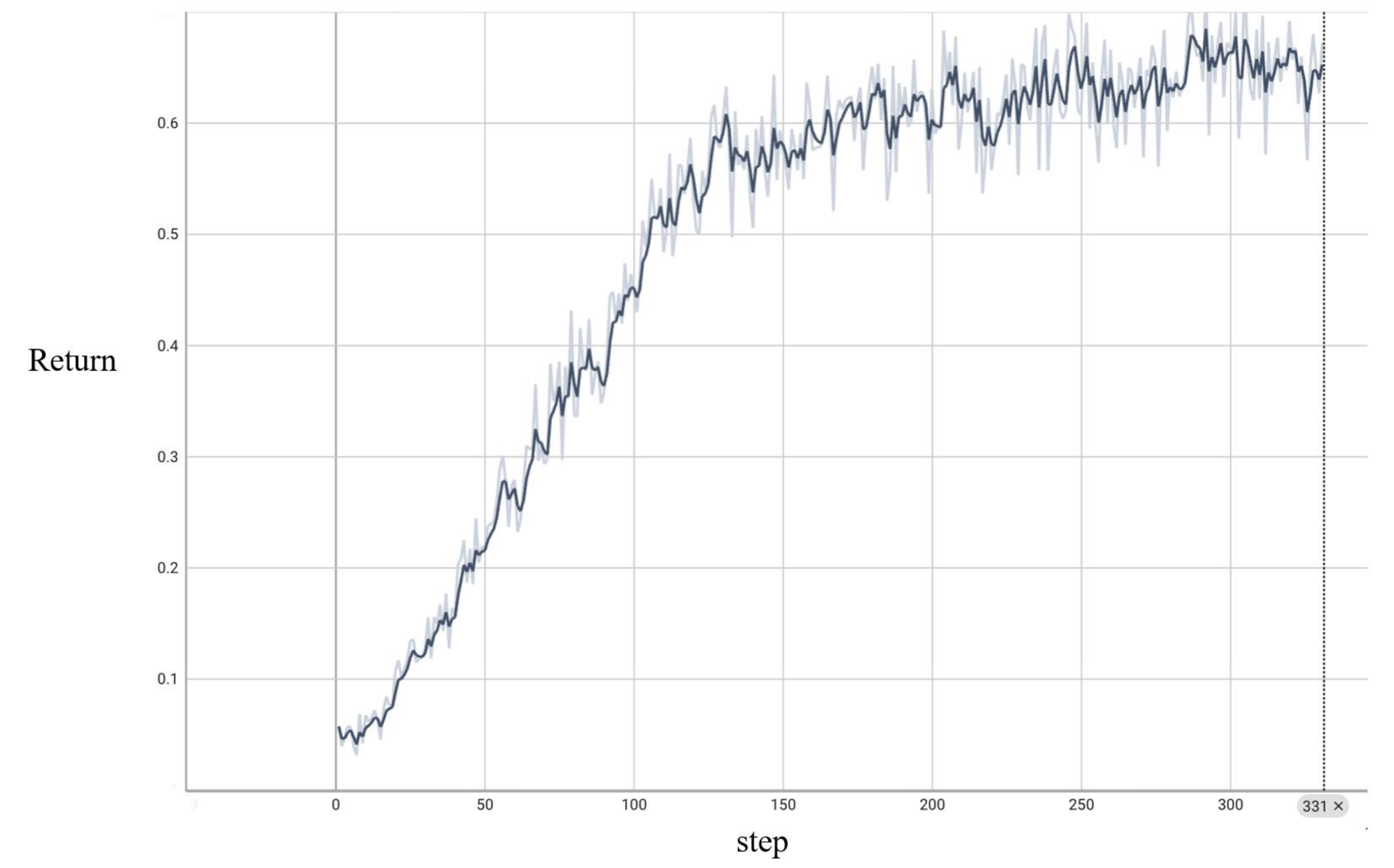}
    \caption{Average returns during training.}
    \label{fig:return_curve}
\end{figure*}

\subsection{Return curves}

Figure \ref{fig:return_curve} illustrates the evolution of average episodic returns throughout the training process, specifically configured with a weighting coefficient of $\beta=0.001$. As demonstrated in the plot, the model exhibits a robust and consistent upward trajectory. The learning curve can be characterized by two distinct phases: an initial phase of rapid ascent, indicating that the agent effectively acquires core competencies and adapts to the task environment; followed by a steady phase of refinement where performance continues to improve before asymptotically stabilizing at a high reward level. This empirical evidence confirms that the chosen hyperparameter setting facilitates reliable convergence towards an optimal policy.

\subsection{Impact of Asymmetric Reward Shaping.} 
\begin{table*}[h]
\caption{\textbf{Ablation study on the reward shaping strategies.} We investigate the impact of different penalty configurations for the judge action. \textit{Symmetric} applies equal penalties ($-0.3$) for both error types. \textit{Recall-Oriented} penalizes missing useful info heavily ($-0.6$) but noise lightly ($-0.3$). Our \textit{Precision-Oriented} strategy (ReSeek) imposes a stricter penalty on noise ($-0.6$) to mitigate error propagation.}
\label{tab:ablation_reward}
\centering
\resizebox{\textwidth}{!}{
\begin{tabular}{lccccccccc}
\toprule[1.25pt]
\textbf{Reward Configuration} & \multicolumn{3}{c}{\textbf{General QA}} & \multicolumn{4}{c}{\textbf{Multi-Hop QA}} & \\
\cmidrule(r){2-4} \cmidrule(r){5-9}
(Reward / FP Penalty / FN Penalty) & \textbf{NQ} & \textbf{TriviaQA} & \textbf{PopQA} & \textbf{HotpotQA} & \textbf{2Wiki} & \textbf{Musique} & \textbf{Bamboogle} & \textbf{FictionalHot} & \textbf{Avg.} \\
\midrule
Symmetric ($+0.3 / -0.3 / -0.3$) & 0.408 & 0.548 & 0.425 & 0.315 & 0.288 & 0.095 & 0.292 & 0.045 & 0.302 \\
Recall-Oriented ($+0.3 / -0.3 / -0.6$) & 0.402 & 0.542 & 0.418 & 0.301 & 0.275 & 0.088 & 0.280 & 0.038 & 0.293 \\
\rowcolor{aliceblue!60}
ReSeek (Ours, $+0.3 / -0.6 / -0.3$) & \textbf{0.415} & \textbf{0.553} & \textbf{0.434} & \textbf{0.328} & \textbf{0.298} & \textbf{0.103} & \textbf{0.304} & \textbf{0.059} & \textbf{0.312} \\
\bottomrule[1.25pt]
\end{tabular}}
\end{table*}
To validate the effectiveness of our asymmetric penalty design, we conducted an ablation study comparing different reward configurations, as shown in Table \ref{tab:ablation_reward}.

We compared our method against two variants: (1) a Symmetric setting, where both false positives (accepting noise) and false negatives (discarding useful info) incur the same penalty of -0.3; and (2) a Recall-Oriented setting, which penalizes missing information more heavily (-0.6) than noise (-0.3).

The results demonstrate that our Precision-Oriented strategy (ReSeek), which imposes a stricter penalty (-0.6) on noise, consistently outperforms other configurations. Notably, the performance gap is most pronounced in multi-hop datasets like HotpotQA and Musique. This indicates that in complex reasoning tasks, preventing the accumulation of irrelevant context (hallucination) is more critical than aggressively recalling every potential piece of evidence. The Recall-Oriented approach, while intuitively appealing for coverage, introduces excessive noise that distracts the LLM, leading to a significant drop in accuracy (Avg. 0.293).

\subsection{Analysis of Structural Adherence and Format Learning}
To ensure the model strictly adheres to the prescribed reasoning structure, we employ a dual-mechanism approach comprising an \textbf{instructional prompt} and an \textbf{automated format enforcement} loop. To verify that the model genuinely learns the structure rather than relying on rejection sampling, we tracked the number of invalid actions requiring retries throughout training.

We tracked the number of format violations at each training step. Given a batch size of 512 and a trajectory length of 4 turns, the model generates a total of 2,048 actions per step. We recorded the count of actions that failed to meet the structural requirements and required correction.

\begin{figure}[h]
    \centering
    \includegraphics[width=0.8\linewidth]{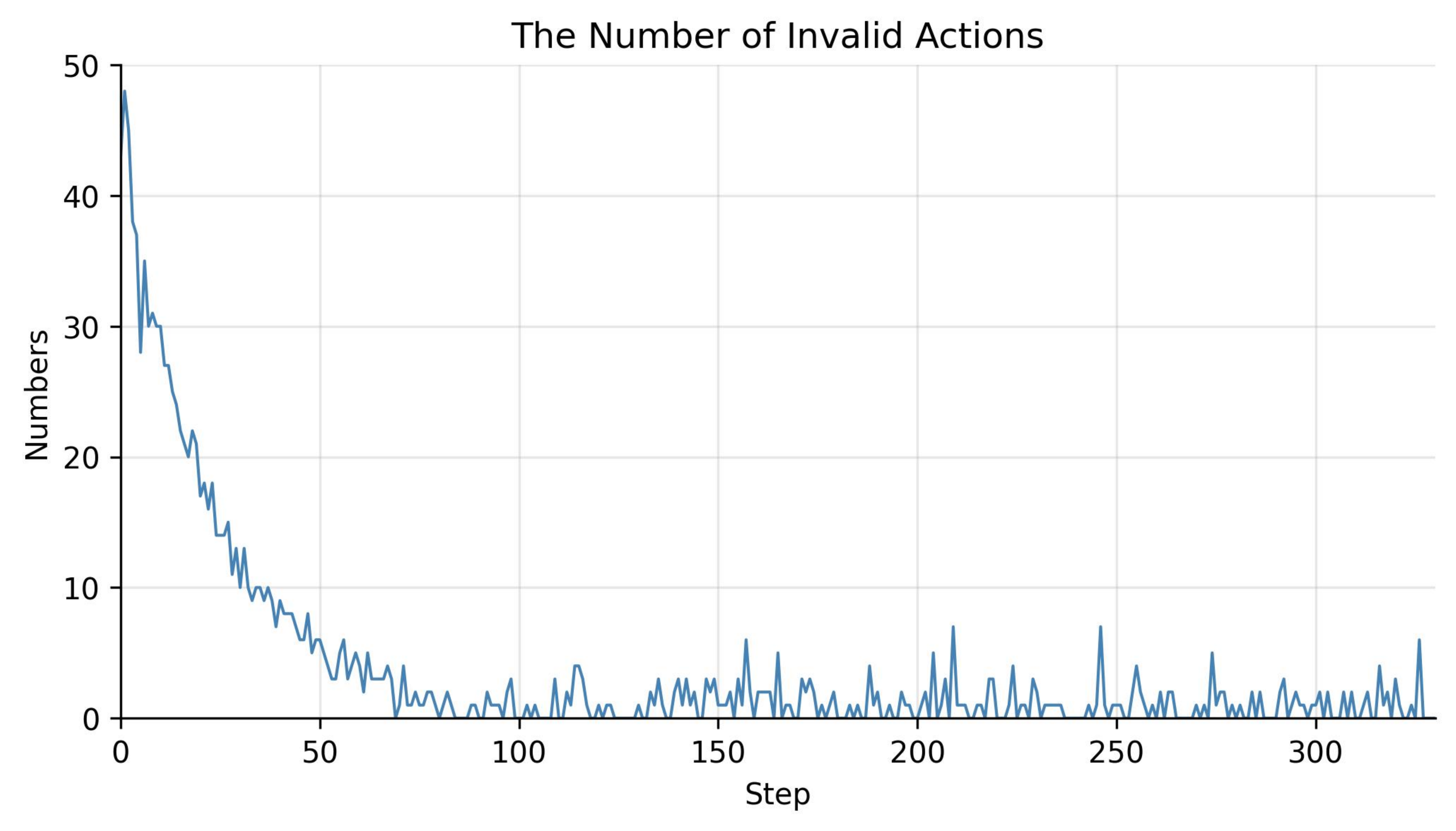}
    \caption{\textbf{Evolution of Invalid Actions During Training.} The plot shows the number of actions requiring format correction per training step (total actions per step = 2048). The sharp drop and subsequent stability at near-zero indicate the model successfully internalizes the structural constraints.}
    \label{fig:structure_curve}
\end{figure}

Figure \ref{fig:structure_curve} illustrates the evolution of invalid action counts over the course of training. As observed, the number of format violations exhibits a peak at the very beginning of training (approximately 50 errors out of 2,048 actions). However, this is followed by a sharp decline within the first 80 steps. Subsequently, the error rate stabilizes at a near-zero level for the remainder of the training duration.

This trend provides strong empirical evidence regarding the model's learning dynamics. The rapid convergence to zero errors indicates that the model is not merely being constrained by the external enforcement mechanism; rather, it is effectively internalizing the desired reasoning structure as an intrinsic part of its policy.

\subsection{Performance-Latency Trade-off Analysis}

\begin{figure}[h]
    \centering
    \includegraphics[width=0.8\linewidth]{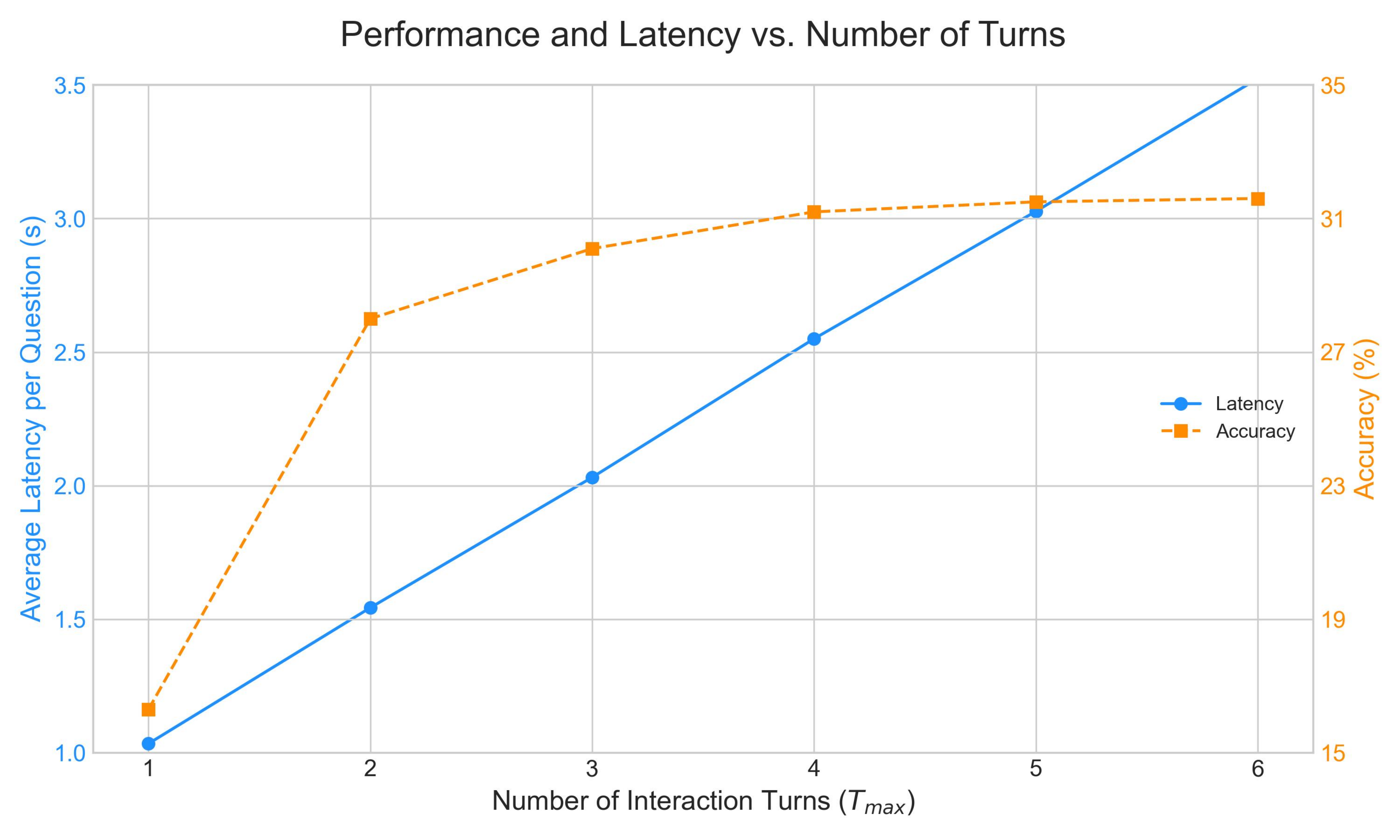}
    \caption{\textbf{Performance vs. Latency Trade-off.} The plot compares model accuracy (orange) and inference latency (blue) across different turn budgets. While latency grows linearly, accuracy gains saturate after $T=4$, justifying our choice of turn budget as the optimal efficiency point.}
    \label{fig:latency_tradeoff}
\end{figure}

To investigate whether the performance improvements stem from genuine reasoning gains rather than mere redundancy, and to justify our choice of turn budget, we analyzed the trade-off between model performance and inference latency across varying numbers of turns.

Figure \ref{fig:latency_tradeoff} visualizes the relationship between accuracy and computational cost. The results reveal two distinct trends:
\begin{itemize}
    \item \textbf{Latency (Blue Line):} As expected, latency increases linearly with each additional turn. This represents a predictable and constant marginal cost for extending the reasoning process.
    \item \textbf{Accuracy (Orange Line):} The model demonstrates substantial performance gains as the turn budget increases up to $T=4$. However, beyond this point, the accuracy plateaus, showing diminishing returns for additional computational expenditure.
\end{itemize}

This saturation point confirms that the multi-turn approach provides genuine reasoning benefits that are maximized at $T=4$. Extending the budget beyond this threshold incurs higher latency without significant performance improvements. Therefore, our selection of $T=4$ strikes an optimal balance, capturing the majority of the reasoning gains while maintaining computational efficiency.

\subsection{Comparison with a recent baseline.}
To address the concern about baseline recency, we compare with HiPRAG~\cite{wu2025hiprag}, a recent method released in Oct 2025. As shown in Table~\ref{tab:hiprag} in the appendix, ReSeek achieves better average performance (42.3 vs.\ 41.5) and leads on NQ, TriviaQA, PopQA, and Musique.
\begin{table}[h]
\caption{\textbf{Comparison with HiPRAG (Qwen2.5-7B).}}
\label{tab:hiprag}
\centering
\small
\setlength{\tabcolsep}{3pt}
\begin{tabular}{lcccccccc}
\toprule
\textbf{Method} & \textbf{NQ} & \textbf{TQA} & \textbf{PopQA} & \textbf{HotQA} & \textbf{2Wiki} & \textbf{Mu} & \textbf{Bam} & \textbf{Avg.} \\
\midrule
HiPRAG & 44.6 & 62.6 & 47.8 & \textbf{45.9} & \textbf{42.8} & 14.0 & 34.8 & 41.5 \\
\rowcolor{aliceblue!60}
ReSeek & \textbf{46.9} & \textbf{64.0} & \textbf{50.1} & 38.9 & 38.2 & \textbf{18.5} & \textbf{39.2} & \textbf{42.3} \\
\bottomrule
\end{tabular}
\end{table}

\begin{table*}[h]
\centering
\caption{\textbf{Detailed performance comparison (Mean $\pm$ Std) and statistical significance tests.} Results are averaged over 5 independent runs.}
\label{tab:main_stats}
\scriptsize
\resizebox{\textwidth}{!}{
\begin{tabular}{lccccccccc}
\toprule
\textbf{Model} & \textbf{NQ} & \textbf{TriviaQA} & \textbf{PopQA} & \textbf{HotpotQA} & \textbf{2wiki} & \textbf{Musique} & \textbf{Bamboogle} & \textbf{FictionalHot} & \textbf{Avg.} \\
\midrule
\multicolumn{10}{l}{\textbf{7B Models}} \\
ZeroSearch & 0.434$\pm$0.005 & \textbf{0.655$\pm$0.006} & 0.490$\pm$0.004 & 0.346$\pm$0.005 & 0.355$\pm$0.006 & 0.184$\pm$0.004 & 0.280$\pm$0.005 & 0.031$\pm$0.004 & 0.346$\pm$0.004 \\
ReSeek & \textbf{0.470$\pm$0.004} & 0.641$\pm$0.006 & \textbf{0.504$\pm$0.005} & \textbf{0.388$\pm$0.007} & \textbf{0.384$\pm$0.005} & \textbf{0.186$\pm$0.003} & \textbf{0.392$\pm$0.003} & \textbf{0.061$\pm$0.002} & \textbf{0.378$\pm$0.004} \\
p-value & 0.005* & 0.985 & 0.038* & 0.003* & 0.012* & 0.471 & 0.001* & 0.004* & 0.009* \\
\midrule
\multicolumn{10}{l}{\textbf{3B Models}} \\
ZeroSearch & 0.410$\pm$0.006 & \textbf{0.574$\pm$0.005} & \textbf{0.448$\pm$0.007} & 0.274$\pm$0.006 & \textbf{0.302$\pm$0.005} & 0.098$\pm$0.004 & 0.111$\pm$0.008 & 0.030$\pm$0.003 & 0.281$\pm$0.004 \\
ReSeek & \textbf{0.414$\pm$0.005} & 0.555$\pm$0.003 & 0.436$\pm$0.006 & \textbf{0.324$\pm$0.008} & 0.300$\pm$0.004 & \textbf{0.104$\pm$0.003} & \textbf{0.305$\pm$0.007} & \textbf{0.059$\pm$0.002} & \textbf{0.312$\pm$0.004} \\
p-value & 0.287 & 0.965 & 0.905 & 0.028* & 0.500 & 0.213 & <0.001* & 0.002* & 0.043* \\
\bottomrule
\end{tabular}}
\end{table*}

\begin{table}[h]
\centering
\caption{\textbf{Ablation on the Number of Turns (Mean $\pm$ Std).}}
\label{tab:ablation_turns_stats}
\begin{tabular}{lccccc}
\toprule
Method & Turn 1 & Turn 2 & Turn 3 & Turn 4 & \textbf{Avg.} \\
\midrule
Search-o1 & 10.0$\pm$0.3 & 18.7$\pm$0.2 & 18.8$\pm$0.3 & 18.3$\pm$0.2 & 16.5$\pm$0.2 \\
Search-R1 & 14.5$\pm$0.2 & 28.8$\pm$0.3 & 28.6$\pm$0.2 & 29.1$\pm$0.3 & 25.3$\pm$0.2 \\
ZeroSearch & 14.0$\pm$0.3 & 28.1$\pm$0.2 & 28.0$\pm$0.3 & 28.2$\pm$0.2 & 24.6$\pm$0.1 \\
\textbf{ReSeek} & 16.3$\pm$0.2 & 28.0$\pm$0.3 & 30.1$\pm$0.2 & \textbf{31.2$\pm$0.1} & \textbf{26.4$\pm$0.2} \\
\bottomrule
\end{tabular}
\end{table}

\begin{table}[h]
\centering
\caption{\textbf{Ablation on Embedding Choice.} (Deterministic values)}
\label{tab:ablation_embed_stats}
\begin{tabular}{lccccc}
\toprule
Backbone & BM25 & E5 & Qwen & Conan & \textbf{Avg.} \\
\midrule
qwen3b-base & 19.1 & 28.8 & 29.0 & 28.8 & 28.9 \\
qwen3b-ins & 21.3 & 31.2 & 31.3 & 31.1 & 31.2 \\
qwen7b-base & 23.5 & 36.1 & 36.2 & 36.0 & 36.1 \\
qwen7b-ins & 25.0 & 37.7 & 38.1 & 37.9 & 37.9 \\
\bottomrule
\end{tabular}
\end{table}

\begin{table*}[h]
\centering
\caption{\textbf{Ablation on Reranker Choice (Mean $\pm$ Std).}}
\label{tab:ablation_reranker_stats}
\resizebox{\textwidth}{!}{
\begin{tabular}{lccccccccc}
\toprule
Method & NQ & TriviaQA & PopQA & HotpotQA & 2Wiki & Musique & Bamboogle & FictionalHot & Avg. \\
\midrule
None & 0.391$\pm$0.004 & 0.495$\pm$0.005 & 0.362$\pm$0.004 & 0.255$\pm$0.006 & 0.218$\pm$0.005 & 0.081$\pm$0.003 & 0.243$\pm$0.004 & 0.025$\pm$0.002 & 0.259$\pm$0.004 \\
Regex & 0.410$\pm$0.003 & 0.541$\pm$0.004 & 0.422$\pm$0.005 & 0.320$\pm$0.004 & 0.291$\pm$0.003 & 0.093$\pm$0.004 & 0.288$\pm$0.005 & 0.042$\pm$0.003 & 0.301$\pm$0.003 \\
Qwen & 0.413$\pm$0.004 & \textbf{0.557$\pm$0.003} & 0.432$\pm$0.003 & 0.326$\pm$0.005 & \textbf{0.301$\pm$0.004} & 0.101$\pm$0.003 & 0.302$\pm$0.004 & 0.057$\pm$0.002 & 0.311$\pm$0.003 \\
\textbf{ReSeek} & \textbf{0.415$\pm$0.003} & 0.553$\pm$0.004 & \textbf{0.434$\pm$0.002} & \textbf{0.328$\pm$0.003} & 0.298$\pm$0.005 & \textbf{0.103$\pm$0.002} & \textbf{0.304$\pm$0.003} & \textbf{0.059$\pm$0.002} & \textbf{0.312$\pm$0.002} \\
\bottomrule
\end{tabular}}
\end{table*}

\subsection{Detailed Experimental Results and Statistics}\label{appendix_stats}
In this section, we provide detailed statistical analyses to substantiate the findings presented in the main text.

\myparagraph{Main Results with Statistical Significance.}
To quantify uncertainty and statistical significance, we expanded our main evaluation (Table~\ref{tab:main}).
First, we report the \textbf{mean and standard deviation} across 5 independent training runs (seeds: 42, 420, 4200, 42000, 420000) for both ReSeek and the strong ZeroSearch baseline.
Second, we conducted a \textbf{paired t-test} comparing ReSeek against Search-R1, reporting exact p-values (* denotes $p < 0.05$).
As shown in Table~\ref{tab:main_stats}, ReSeek demonstrates statistically significant improvements across the majority of benchmarks.

\myparagraph{Robustness of Ablation Studies.}
We provide variance analysis for our ablation studies.
Table~\ref{tab:ablation_turns_stats} shows the turn ablation with std, confirming robust gains.
Table~\ref{tab:ablation_embed_stats} lists the exact values for the embedding ablation (deterministic given fixed seed).

\subsection{Case study}\label{appendix_case study}
To provide a concrete illustration of ReSeek's advantages, we present a side-by-side case study. We compare the reasoning process of ReSeek against Search-R1 on some multi-hop questions. 

\subsubsection{Case study I}
\begin{figure*}[h]
\begin{center}
\vskip -0.1in
\centerline{\includegraphics[width=1.0\columnwidth]{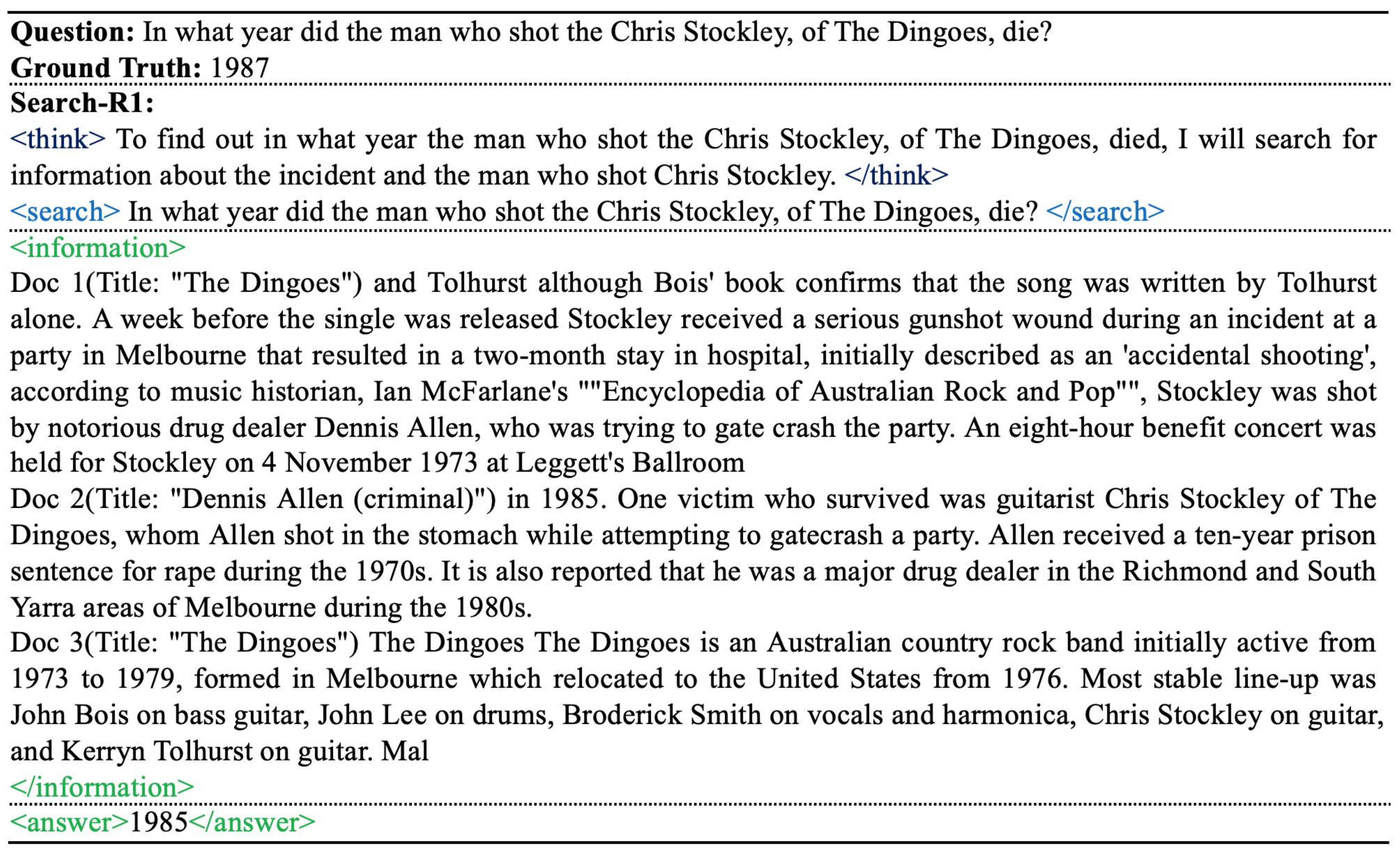}}
\vskip -0.1in
\caption{\textbf{A baseline agent (Search-R1) failing the two-hop question.} The agent attempts to solve the problem in a single step and incorrectly extracts the year of the shooting (1985) instead of the correct year of death (1987).}
\label{fig:r1_case1}
\end{center}
\vskip -0.1in
\end{figure*}
\begin{figure*}[h]
\begin{center}
\centerline{\includegraphics[width=1.0\columnwidth]{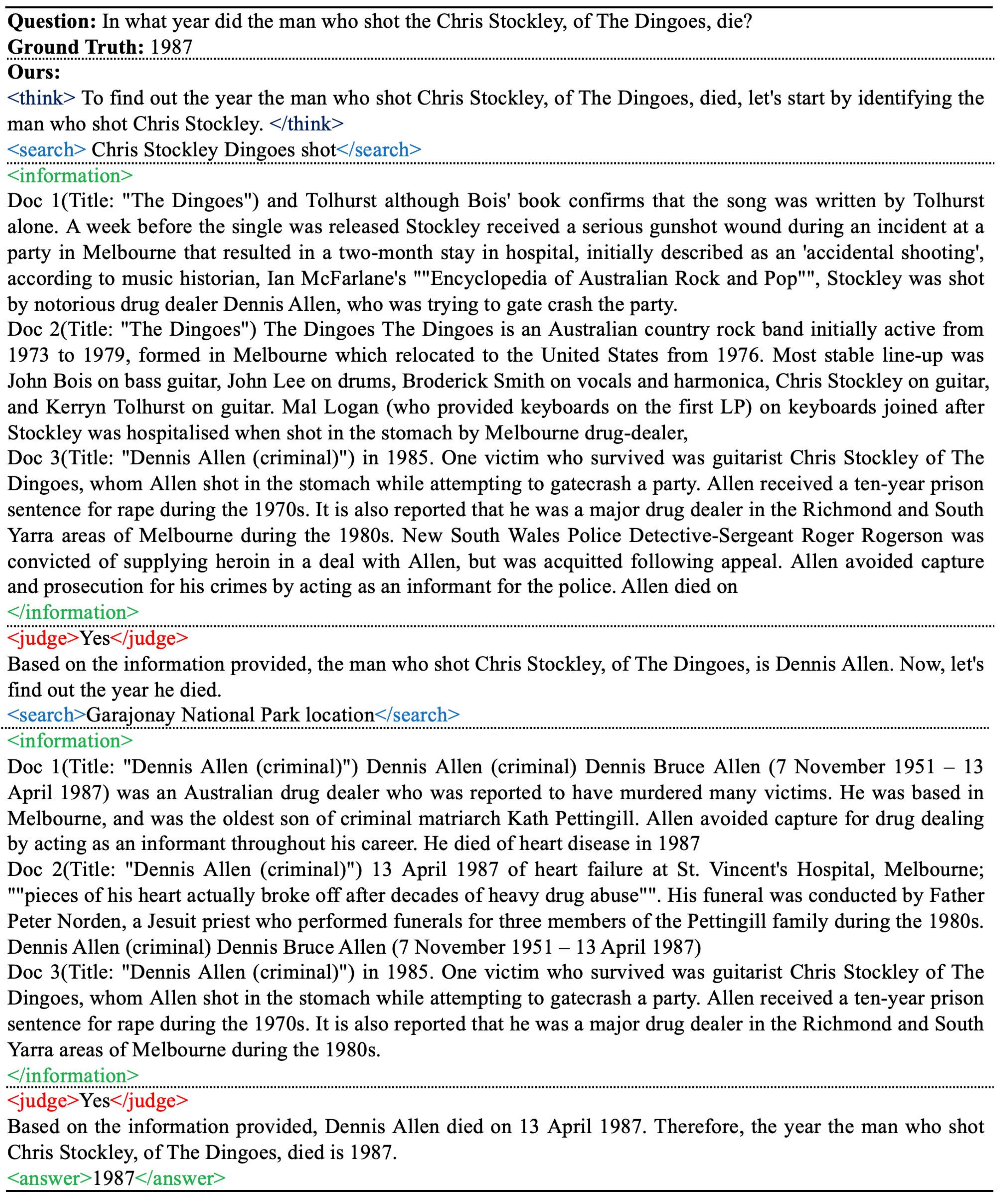}}
\caption{\textbf{A case study of ReSeek on a two-hop question.} The agent first identifies the shooter (``Dennis Allen") and then finds his death year. The judge action is used to validate the intermediate finding before proceeding to the second reasoning step.}
\label{fig:reseek_case1}
\end{center}
\end{figure*}

This side-by-side comparison highlights the value of ReSeek's self-correction. The baseline agent (Figure\ref{fig:r1_case1}), treating the query as a single task, incorrectly extracts a distractor date (1985, the year of the shooting) and fails. 
In contrast, ReSeek's trajectory (Figure\ref{fig:reseek_case1}) demonstrates robust multi-step reasoning. It first identifies the shooter (``Dennis Allen"), then uses the judge action to validate this intermediate result before initiating a second, focused search for the death year. This structured process prevents premature conclusions and leads to the correct answer where the baseline fails.

\subsubsection{Case study II}\label{appendix_case2}

\begin{figure*}[h]
\begin{center}
\vskip -0.1in
\centerline{\includegraphics[width=1.0\columnwidth]{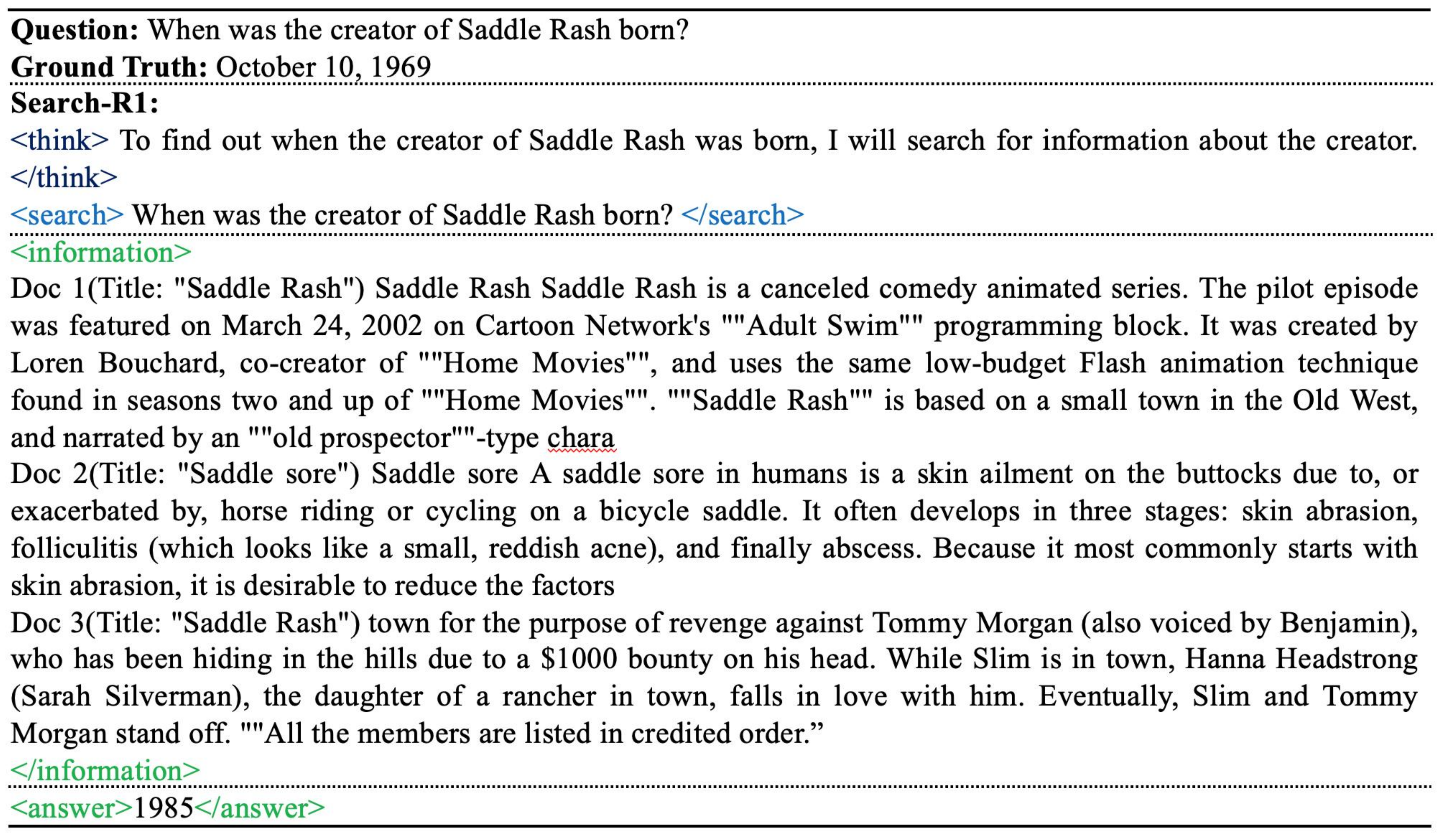}}
\vskip -0.1in
\caption{\textbf{A baseline agent (Search-R1) failing the two-hop question.} While the agent's search successfully identifies the creator, ``Loren Bouchard," it fails to perform the necessary follow-up search for their birth date. It prematurely concludes with a hallucinated and incorrect answer.
}
\label{fig:r1_case2}
\end{center}
\vskip -0.1in
\end{figure*}

\begin{figure*}[h]
\begin{center}
\centerline{\includegraphics[width=1.0\columnwidth]{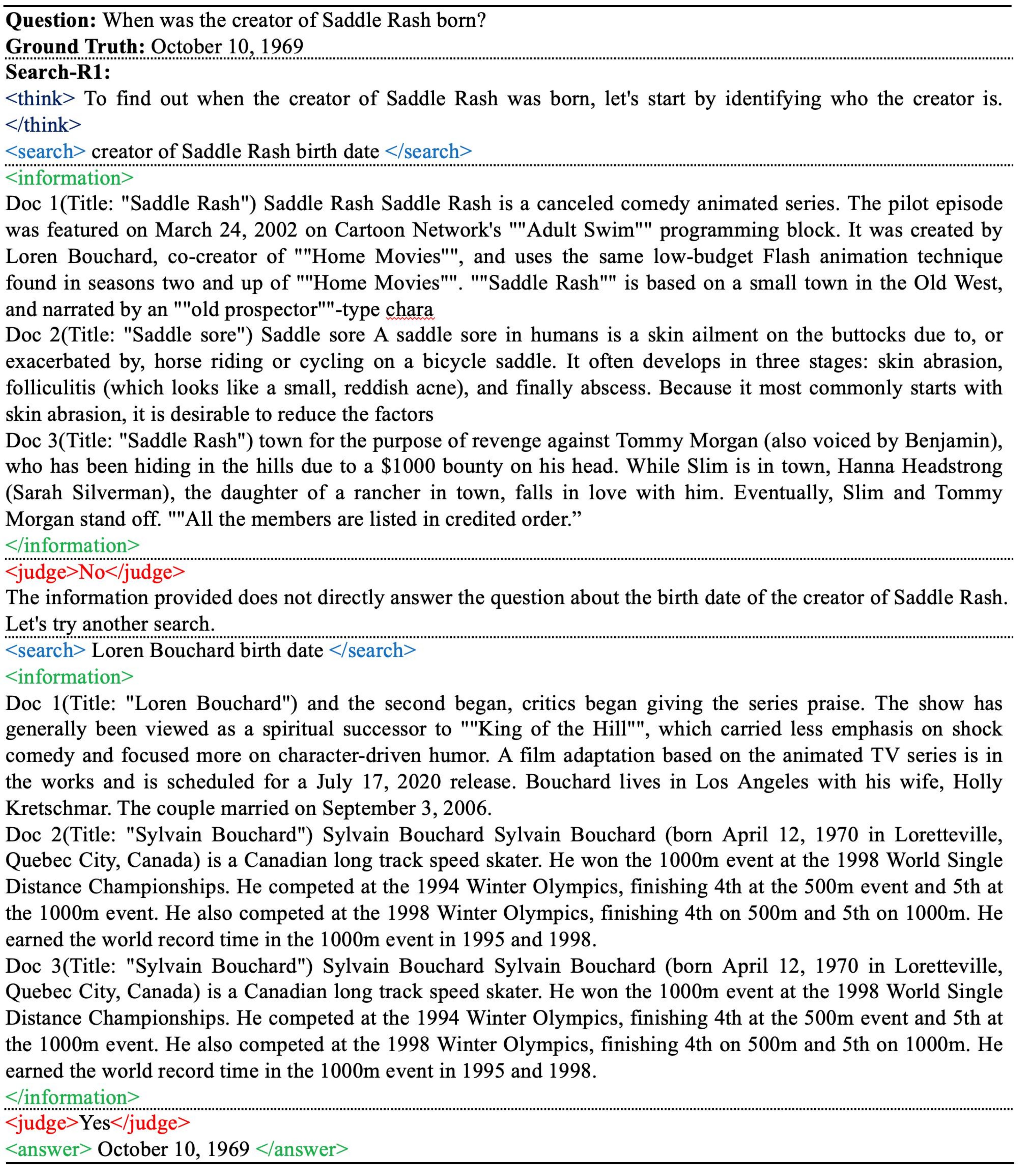}}
\caption{\textbf{A case study of ReSeek on a two-hop question.} The agent first attempts a broad search but correctly uses the judge action to determine the retrieved information is insufficient. It then extracts the creator's name (``Loren Bouchard") from the initial context and initiates a second, focused search for the birth date.}
\label{fig:reseek_case2}
\end{center}
\end{figure*}
The contrast between ReSeek's success (Figure\ref{fig:reseek_case2}) and the baseline's failure (Figure\ref{fig:r1_case2}) on this second case study is even more telling. The baseline agent correctly identifies the creator's name, ``Loren Bouchard," but its monolithic reasoning process stops there. Unable to find the birth date in the initial context and lacking a mechanism to initiate a follow-up query, it resorts to hallucinating an incorrect answer (``1985") that is entirely absent from the evidence.

In stark contrast, ReSeek demonstrates the power of structured self-correction. After its initial search, the judge action correctly identifies that the answer has not been found (\judge{No}No). This crucial validation step prompts the agent to formulate a new plan: use the newly found entity, ``Loren Bouchard," to perform a second, targeted search. This methodical decomposition of the problem allows ReSeek to navigate the multi-hop query, successfully retrieve the correct birth date, and avoid the pitfall of ungrounded generation that caused the baseline to fail.

\subsection{Failure case study}\label{appendix_failure_case}

\begin{figure*}[h]
\begin{center}
\centerline{\includegraphics[width=1.0\columnwidth]{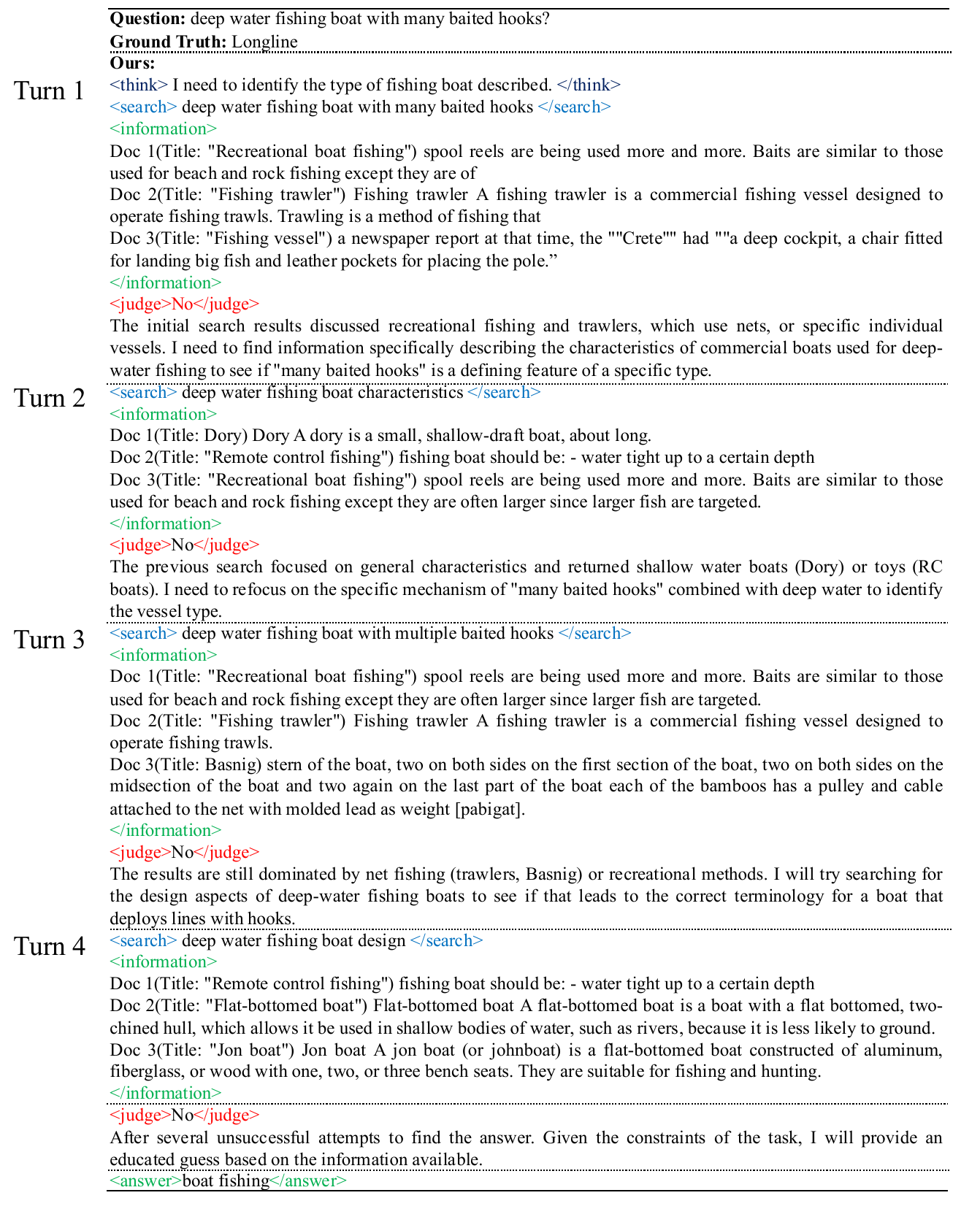}}
\caption{A case study illustrating a retrieval failure. The model attempts to identify the specific term ``Longline" based on its functional description. Despite performing four rounds of iterative search and query reformulation (shifting focus from the full question to specific ``characteristics" and ``design"), the retrieval system consistently returns irrelevant documents regarding net-based fishing (e.g., ``Trawler") or shallow-water vessels (e.g., ``Dory," ``Jon boat"). Consequently, lacking the necessary evidence in the retrieved context, the model is unable to derive the correct entity and resorts to an incorrect guess.}
\label{fig:reseek_case3}
\end{center}
\end{figure*}

To further investigate the concern regarding reward hacking and understand the performance plateau at higher turn limits, we analyze a representative failure case shown in Figure~\ref{fig:reseek_case3}. In this episode, the user asked for a ``deep water fishing boat with many baited hooks'' (Ground Truth: \textit{Longline fishing}). The maximum turn limit ($T_{max}$) was set to 10. As illustrated in the figure, the agent actively reformulated its search queries four times but failed to retrieve the target document, eventually providing an incorrect answer. Crucially, despite having a remaining budget of 6 turns, the agent voluntarily terminated the process at Turn 4.

This case highlights two key observations:

\begin{itemize}
    \item \textbf{Corpus Limitations (Reason for Failure):} The failure was not due to a lack of effort or poor reasoning by the agent. The agent demonstrated effective \textit{re-planning} capabilities, shifting its search strategy from the full description to specific keywords like ``characteristics,'' ``multiple baited hooks,'' and ``design.'' However, the retrieval system consistently returned irrelevant documents (e.g., \textit{Trawlers}, \textit{Dories}, and \textit{Basnigs}), and the specific document for ``Longline fishing'' was not retrieved. This confirms that the performance plateau observed in our main experiments is primarily driven by information retrieval limits in the Wiki-18 corpus rather than agent inefficiency.
    
    \item \textbf{Evidence Against Reward Hacking:} This trajectory serves as strong empirical evidence that \textit{reward hacking does not occur}. If the agent were motivated to artificially prolong the episode to ``farm'' rewards, it would be optimal to continue generating actions until Turn 10. Instead, the agent's decision to stop early indicates that its learned policy correctly balances the potential gain of finding the answer against the penalty of additional steps. As seen in the final reasoning step (``I have exhausted reasonable search queries''), when the agent determines that further search is futile, it terminates the episode immediately to minimize the penalty.
\end{itemize}

\end{document}